\documentclass[12pt,letterpaper]{article}

\usepackage{natbib,hyperref}

\usepackage[ left=1in, top=1in, right=1in, bottom=1in]{geometry}
\usepackage{graphicx,bm,colonequals,amsmath,amssymb,url,xcolor,bbm}
\usepackage{array,tabularx,multirow}
\usepackage[font={footnotesize}]{caption,subcaption}
\usepackage[utf8]{inputenc}
\usepackage{enumitem}
\usepackage{soul} \usepackage{placeins} \usepackage[normalem]{ulem}
\usepackage{booktabs}
\usepackage{adjustbox}

\graphicspath{{figures/}{contents/}{plots/}}

\usepackage{xr}
\makeatletter
\newcommand*{\addFileDependency}[1]{  \typeout{(#1)}
  \@addtofilelist{#1}
  \IfFileExists{#1}{}{\typeout{No file #1.}}
}
\makeatother
\newcommand*{\myexternaldocument}[1]{    \externaldocument{#1}    \addFileDependency{#1.tex}    \addFileDependency{#1.aux}}
\myexternaldocument{supplement}

\usepackage{mathtools}
\mathtoolsset{showonlyrefs}

\everydisplay{\textstyle}

\bibpunct[, ]{(}{)}{;}{a}{,}{,}

\usepackage{amsthm}
\newtheoremstyle{propstyle}     {2mm}                        {1mm}                        {\itshape}                       {}                               {\scshape}                       {.}                              {.5em}                           {}  \theoremstyle{propstyle}

\theoremstyle{propstyle}

\theoremstyle{propstyle}

\theoremstyle{propstyle}

\theoremstyle{propstyle}

\usepackage[ruled,vlined]{algorithm2e}
\SetKwInput{KwInput}{Input}
\usepackage{algorithmic}
\usepackage{array,framed}
\usepackage{float}

\makeatletter
\renewcommand{\paragraph}{  \@startsection{paragraph}{4}  {\z@}{2ex \@plus 1ex \@minus .2ex}{-1em}  {\normalfont\normalsize\bfseries}}
\makeatother

\DeclareMathAlphabet\mathbfcal{OMS}{cmsy}{b}{n}

\renewcommand{\bf}{\mathbf{f}}
\newcommand{\bh}{\mathbf{h}}
\newcommand{\bu}{\mathbf{u}}

\newcommand{\ba}{\mathbf{a}}

\newcommand{\bb}{\mathbf{b}}
\newcommand{\bs}{\mathbf{s}}

\newcommand{\bx}{\mathbf{x}}
\newcommand{\by}{\mathbf{y}}

\newcommand{\bS}{\mathbf{S}}

\newcommand{\bF}{\mathbf{F}}

\newcommand{\bH}{\mathbf{H}}

\newcommand{\bfzero}{\mathbf{0}}

\newcommand{\bfmu}{\bm{\mu}}

\newcommand{\bftheta}{\bm{\theta}}

\newcommand{\bfdelta}{\bm{\delta}}

\newcommand{\bfbeta}{\bm{\beta}}
\newcommand{\bfepsilon}{\bm{\epsilon}}

\newcommand{\bfSigma}{\bm{\Sigma}}

\newcommand{\normal}{\mathcal{N}}

\title{Amortized Probabilistic Retrieval of Atmospheric CO$_2$ from OCO-2 Spectra Using Deep Learning with Laplace Approximations and Normalizing Flows}

\author{Alejandro Calle-Saldarriaga\thanks{Department of Statistics, University of Wisconsin--Madison} \and Felix Jimenez\footnotemark[1]  \and Jack Grosskreuz\footnotemark[1] \and Jiazheng Wang\footnotemark[1]  \and Jonathan Hobbs\thanks{Jet Propulsion Laboratory, California Institute of Technology} \and Matthias Katzfuss\footnotemark[1] \thanks{Corresponding author: \texttt{katzfuss@gmail.com}}}

\date{}

\usepackage{cleveref}

\begin{document}

\maketitle

\begin{abstract}
Space-based monitoring of atmospheric carbon dioxide (CO$_2$) constrains the global carbon budget. NASA's Orbiting Carbon Observatory-2 (OCO-2) estimates column-averaged dry-air mole fractions of CO$_2$ (XCO$_2$) from high-resolution spectra, but operational retrievals are computationally expensive and impose stringent Gaussianity assumptions on the retrieved posterior. We present a deep learning framework that addresses both through amortized probabilistic inference. Lacking ground truth for real observations, we train and evaluate on a high-fidelity OCO-2 simulation ensemble with calibrated forward-model errors, comparing against the version-10 operational full-physics retrieval, without its bias correction, on the same radiances. Our architecture encodes each spectral band separately and estimates posteriors of the full CO$_2$ column, or summaries thereof, with Laplace approximations and conditional normalizing flows. Once trained, inference costs milliseconds per sounding rather than minutes. Trained on simulations that explicitly include forward-model discrepancy, our retrievals are more accurate than the operational one for XCO$_2$ on both data partitions we consider, and competitive on profiles. The flow represents asymmetric posteriors that a Gaussian cannot, a gain attributable to shape rather than scale, and its advantage in predictive density persists where its accuracy advantage does not. These results are established on a land-only ensemble against one configuration of the operational algorithm. On held-out reference soundings and two unseen months the XCO$_2$ and density advantages persist. For XCO$_2$, coverage is near nominal on February data and, though below nominal on the unseen months, remains far closer to it than the operational retrieval's.

\end{abstract}

\section{Introduction}
\label{sec:intro}

The accumulation of greenhouse gases, particularly carbon dioxide ($\rm{CO}_2$), is a primary driver of anthropogenic climate change \citep{arrhenius1896xxxi, ipcc_ar6_syr_2023}. To mitigate these effects and move towards net-zero emissions, it is essential to accurately estimate the global geographic distributions of carbon sources and sinks. Space-based remote sensing offers the only viable pathway for obtaining these measurements on a global scale with high spatiotemporal resolution. Missions such as NASA's Orbiting Carbon Observatory-2 and 3 (OCO-2/3) \citep{eldering2017orbiting, basilio2019orbiting, taylor2020} measure sunlight reflected from the Earth's surface within specific spectral bands. These radiance measurements are processed to infer the column-averaged dry-air mole fraction of $\rm{CO}_2$, denoted as XCO$_2$. This inverse problem is termed a {\it{retrieval}}.

Precise estimates of XCO$_2$ are critical for ``top-down'' atmospheric inversion models that quantify surface fluxes \citep{miller_precision_2007}. The tolerances are demanding: systematic retrieval errors below one part per million (ppm) are enough to alter the flux estimates \citep{chevallier2019}. They are also not comfortably met: in inverse-model intercomparisons the annual global flux is recovered to within roughly $20\%$, but agreement degrades quickly at subcontinental scales, where estimates can differ by more than $100\%$ of the annual flux, and the contributions of the retrieval schemes and of the inverse models to that spread are of comparable size \citep{houweling2015}; the same contrast between global agreement and regional disagreement is found in the OCO-2 era \citep{crowell2019}. OCO-2 retrievals are also used to attribute interannual variability in the carbon cycle to particular regions and processes, such as the role of tropical Pacific outgassing during the 2015--2016 El Ni\~{n}o \citep{chatterjee2017}, and to detect and quantify localized sources, from megacities and volcanoes \citep{schwander2017} to individual power plants \citep{nassar2021}. What each of these applications consumes is not the retrieved value alone but the uncertainty attached to it: it sets the weight a sounding receives in an inversion, and it decides whether an apparent enhancement is reported as a detection. Inversions moreover ingest the retrieval through its averaging kernel and pressure weighting rather than as a bare column mean, so the vertical structure of the estimate matters as well as its aggregate, and the near-surface part of the column is the part most directly connected to surface exchange. For these inversions to be reliable, retrievals must be not only accurate but also accompanied by rigorous uncertainty quantification (UQ). Furthermore, the sheer volume of data produced by modern hyperspectral satellites presents a massive computational challenge for traditional processing pipelines.

The retrieval of atmospheric state variables from observed satellite radiances is a nonlinear inverse problem \citep{rodgers2000inverse}. Formally, let $\mathbf{x}$ represent the geophysical state vector (including atmospheric $\rm{CO}_2$ profiles, surface pressure, aerosols, and albedo) and $\mathbf{y}$ represent the measured spectral radiances. The physical relationship is modeled by a forward model $\mathbf{F}$ such that:
\begin{equation}\label{eq:inv_problem}
    \mathbf{y} = \mathbf{F}(\mathbf{x}, \mathbf{b}) + \boldsymbol{\delta} + \boldsymbol{\epsilon},
\end{equation}
where $\mathbf{b}$ denotes ancillary parameters, $\boldsymbol{\epsilon}$ represents instrument noise derived from the instrument's signal-to-noise ratio., and $\boldsymbol{\delta}$ represents systematic model discrepancy—the inevitable structural error where the physics-based forward model fails to perfectly replicate reality \citep{braverman2021post}.

Current operational retrievals, such as the Atmospheric Carbon Observations from Space (ACOS) ``full-physics'' algorithm \citep{odell2012ACOS, crisp2012ACOS}, rely on optimal estimation (OE) techniques \citep{rodgers2000inverse}. While rigorous, these approaches face three significant limitations. First, they are computationally expensive, requiring iterative radiative transfer calculations for every single sounding, which creates bottlenecks for real-time global monitoring. Second, standard OE assumes Gaussian posteriors, effectively linearizing the problem near the solution. This assumption often breaks down in the presence of nonlinear physical effects (e.g., cloud and aerosol scattering) and misspecification of the a priori state \citep{nguyen2019,patil2022objective}, leading to poor uncertainty estimates \citep{connor2016UQ, hobbs2017simulation}. Third, operational retrievals often assume the forward model is perfect ($\boldsymbol{\delta} = \bfzero$), whereas extensive validation studies have shown that forward model error is a dominant source of bias in satellite retrievals \citep{kulawik2019validation}.

To address the computational bottleneck, recent research has explored machine learning (ML) and deep learning for atmospheric retrievals. Approaches using neural networks \citep{david2021xco, breon2022potential}, transformers \citep{chen2025transformer}, and nonlinear embeddings \citep{keely2023nonlinear} have demonstrated the ability to approximate physical solvers with high accuracy. However, purely data-driven approaches raise a question about the provenance of the predictive signal; for instance \citet{breon2022potential} noted that networks trained on real data may over-rely on ancillary correlations (e.g., latitude or date) to mimic climatology, rather than extracting the signal from the spectral absorption features. Furthermore, many existing ML methods focus primarily on point estimates or provide simple Gaussian uncertainty outputs that inherit the limitations of the training data \citep{chen2025deterministic}. \citet{keely2025conditional} combines the speed of deep learning with probabilistic modeling that captures possibly non-Gaussian posterior distribution for XCO$_2$ via diffusion models, however they only propose and evaluate the method for this scalar prediction task; there remains a need for a framework that can leverage deep learning for prediction of the highly correlated joint posterior of the full $\rm{CO}_2$ column (or subsets of the full column). A large body of work applies ML downstream of the retrieval, producing gap-free XCO$_2$ fields from the retrieved product via geostatistical interpolation \citep{hammerling2012mapping}, and more recently via gradient-boosted and deep models trained on the retrievals together with meteorological and land-surface covariates, at regional \citep{he2023machine, liu2024mapping} and global \citep{antezanalopez2025global, hwang2026reconstructing} scale. These methods take the retrievals as input and address its coverage, while this work takes radiances as input and produces the retrieval itself.  The computational bottleneck could also be addressed in alternative ways, for example \citet{lamminpaa2025} leaves the retrieval intact and emulates the radiative-transfer forward model so that the iterative solve becomes affordable.

There remains a need for a framework that combines the speed of deep learning with flexible, rigorous probabilistic modeling and captures complex, non-Gaussian posterior distributions. We note that probabilistic modeling does not by itself resolve the question of where the predictive signal comes from; we address that question directly in Section \ref{sec:occlusion}.

In this work, we present an amortized probabilistic retrieval framework validated using a high-fidelity OCO-2 simulation dataset \citep{braverman2021post}, while noting that extending it to OCO-3 spectra is straightforward: they share the same spectrometer and are processed via the same retrieval algorithm; what needs to be redone is the simulation ensemble to capture possibly new viewing geometries \citep{basilio2019orbiting, taylor2020}. While the Total Carbon Column Observing Network (TCCON) \citep{wunch2011, laughner2024total} is the standard for XCO$_2$ validation, it is insufficient for the rigorous evaluation of multidimensional probabilistic profiles. TCCON sites are geographically sparse, and satellite-to-ground comparisons are confounded by co-location errors and atmospheric-state filter choices that introduce significant secondary uncertainties \citep{li2023influence, yadav2025error}. Furthermore, reconciling disparate vertical sensitivities and pressure grids requires averaging-kernel corrections and interpolations that further obscure the comparison \citep{nguyen2020,mendonca2021assessing}. Critically, TCCON lacks the joint posterior distribution data required to benchmark the high-dimensional UQ produced by our Laplace and normalizing flow methods. We therefore utilize a high-fidelity simulation environment that explicitly incorporates quantified forward model errors \citep{hobbs2020spectroscopic}. This provides an absolute, joint ground truth across the full 20-layer state vector, enabling a controlled assessment of our model's ability to capture complex atmospheric states and non-Gaussian uncertainties. The ensemble is land-only and drawn from three months of 2020, so the assessment is bounded by the observing conditions it samples, a limitation we return to in Sections \ref{subsec:group_split} and \ref{sec:conclusions}.

The specific contributions of this paper are as follows:
\begin{enumerate}
    \item \textbf{Amortized inference:} We demonstrate that our deep learning approach is orders of magnitude faster at test time than operational iterative solvers, at a one-time training cost that is recovered within a few minutes of instrument data.
    \item \textbf{Robustness to model error:} By training on simulations that include calibrated forward-model discrepancies, estimated by matching the distribution of operational OCO-2 spectral residuals, our networks learn to account for systematic errors neglected by standard inversions.
    \item \textbf{High-accuracy point estimates:} Our predictions for held-out simulated
XCO$_2$ are more accurate than the raw estimates of traditional OE, on both of the
partitions we consider; for the full $\rm{CO}_2$ column the two are comparable, with the
learned models ahead within the geophysical regimes the reference set samples densely. With its mean error removed, the OE estimate of XCO$_2$ remains less accurate within reference but is close to ours on unseen reference soundings (Section~(Section~\ref{subsec:group_split}).
    \item \textbf{UQ with Laplace and normalizing flows:} We provide rigorous UQ, utilizing both Laplace approximations and Normalizing Flows, that is empirically better calibrated than the full-physics retrieval run on the same radiances for XCO$_2$ in every setting we consider and for the $20$-level profile on both February partitions; for near-surface levels the comparison depends on the method and setting (Section~\ref{subsec:group_split}).
    \item \textbf{Non-Gaussian posteriors:} Through the use of normalizing flows, we successfully model asymmetric posterior distributions, overcoming the restrictive Gaussian assumptions of traditional retrieval theory.
\end{enumerate}

The remainder of this paper is organized as follows. Section \ref{sec:data} describes the high-fidelity simulation dataset and the forward-model error characterization. Section \ref{sec:method} details our neural network architecture and the implementation of Laplace approximations and normalizing flows. Section \ref{sec:experiments} presents the experimental setup and results, focusing on accuracy, calibration, and computational efficiency. Finally, Section \ref{sec:conclusions} offers concluding remarks and directions for future work. Code for reproducing results described in this paper can be found in \url{https://github.com/acallesalda/oco2-amortized-uq}, archival code repository in \\url{https://zenodo.org/records/24025972}, and data and trained model weights can be found at \url{https://zenodo.org/records/24025972}.

\section{Data: High-fidelity OCO-2 simulations}
\label{sec:data}

To rigorously validate the calibration of our UQ and the robustness of our method to structural errors, we utilize a high-fidelity simulation dataset developed specifically for the OCO-2 mission. While real observational data is abundant, it lacks the dense, absolute ``ground truth'' required to verify probabilistic coverage and small-scale accuracy. The dataset used in this work \citep{braverman2021post, hobbs2020spectroscopic} bridges this gap by generating synthetic radiances that simulate not only the physics of radiative transfer but also the systematic imperfections of the forward models used in operational retrievals.

\subsection{The simulation framework}
\label{subsec:sim_framework}

The data-generation process follows the standard remote-sensing forward problem but introduces a critical component: model discrepancy. Let $\mathbf{x}$ be a high-dimensional state vector containing approximately 50 elements, including the profile of atmospheric $\rm{CO}_2$ (from which XCO$_2$ is derived), surface pressure, water vapor, temperature adjustments, aerosol properties, surface albedo, and instrument parameters.

The simulated radiance vector $\mathbf{y}$ is generated via a physical forward model $\mathbf{F}(\mathbf{x}; \mathbf{b})$, where $\mathbf{b}$ represents fixed ancillary parameters. To create a realistic testing ground that mimics the challenge of real-world retrievals, the dataset includes both measurement noise and model discrepancy (Equation~\eqref{eq:inv_problem}).

Crucially, $\boldsymbol{\delta}$ represents \textit{model discrepancy}: the error the forward model makes at the true state. Letting $F_1$ be the operational full-physics forward model and $F_0$ the mapping that nature realizes, $\bfdelta = F_0(\bx; \bb_0) - F_1(\bx; \bb_1)$, with the same state in both terms. It is a vector in radiance space which reflects known limitations in absorption coefficients, aerosol scattering approximations, and spectroscopic uncertainties \citep{braverman2021post,hobbs2020spectroscopic}. In operational contexts, the inverse solver assumes $F_0(\bx; \bb_0) = F_1(\bx; \bb_1)$, i.e., $\bfdelta = \bfzero$.

 In the simulation used in this work, $\bfdelta \sim \mathcal{N}(\tilde \bfmu_{c}, \tilde \bfSigma_{c})$, where $c = 1, \dots, 9$ indexes the geophysical clusters described in Section \ref{subsec:dataset_comp}. The residuals of operational OCO-2 retrievals reflect the
combined effect of $\bfdelta$ and $\bfepsilon$, those of a discrepancy-free simulation
reflect $\bfepsilon$ alone, and matching the first two moments of the two ensembles
identifies $\tilde\bfmu_{c}$ and $\tilde\bfSigma_{c}$. The discrepancy
distribution is therefore anchored to real observations rather than to perturbations of
the state.

By training our deep learning models on data that include $\boldsymbol{\delta}$, we allow the network to implicitly learn to correct or marginalize over these systematic errors. This is a capability that standard  ``full-physics'' solvers, as usually written, lack (this will be introduced in Section \ref{subsec:baselines}). We note however that if one could estimate a discrepancy mean and covariance matrix over spectral space, perhaps via the same procedure outlined in this subsection, one could include such terms in a OE cost function and thus model this discrepancy.

\subsection{Dataset composition and inputs}
\label{subsec:dataset_comp}

The dataset is structured around distinct geophysical regimes to ensure global representativeness. It contains 9 distinct ``clusters'' representing different observing conditions and geophysical states (e.g. high vs.\ low latitude, varying surface albedos and aerosol conditions) over land. These clusters were assembled using self-organizing maps (SOM) on an ensemble of operational OCO-2 retrievals \citep{v10dug}. Within each cluster, two distributions are estimated separately: An ensemble distribution for the states $\bx$, fitted as a Gaussian mixture to the retrieved state vectors of that cluster, and the discrepancy distribution $\bfdelta \sim \mathcal{N}(\tilde \bfmu_{c}, \tilde \bfSigma_{c})$. The state ensemble is therefore not Gaussian, which bears directly on the posterior geometry discussed in Section \ref{sec:nongauss}. The discrepancy distribution is thus indexed by cluster, and has constant moments within the clusters, which are sets of soundings with similar geophysical and observing conditions (as opposed to contiguous regions).

The SOM was trained on retrieved XCO$_2$, solar-induced fluorescence, surface pressure, total aerosol optical depth (AOD) and weak CO$_2$ band surface reflectance, each centered and scaled globally beforehand. Reference soundings are allocated in proportion to cluster size and drawn by a Latin hypercube over solar zenith angle, footprint index and prior AOD; the $100$ replicates per sounding use variability and correlation structures estimated at the cluster level.

Because the discrepancy enters additively and is drawn independently of the state and of
the instrument noise, its effect on the distribution of simulated radiances within a
cluster follows directly: $\mathbb{E}[\by] = \mathbb{E}[\bF(\bx;\bb)] +
\tilde\bfmu_{c}$ and $\mathrm{Cov}[\by] = \mathrm{Cov}[\bF(\bx;\bb)] +
\tilde\bfSigma_{c} + \bS_{\bfepsilon}$. The discrepancy therefore shifts the mean of
the radiance distribution and alters both the per-channel variances and the correlation
structure across channels, while leaving the distribution of the states themselves
unchanged.

Although $\bfdelta$ is high-dimensional, $\tilde\bfSigma_{c}$ is never formed
densely. It is estimated by the moment-matching procedure described in Section
\ref{subsec:sim_framework} and then projected onto its leading principal components, which
both guarantees positive-definiteness and keeps sampling inexpensive; the effective
dimension of the sampled discrepancy is far below the number of spectral channels.

\begin{figure}[htpb]
    \centering
    \includegraphics[
        width=\linewidth,
        trim={0cm 2.5cm 0cm 2.5cm},
        clip
    ]{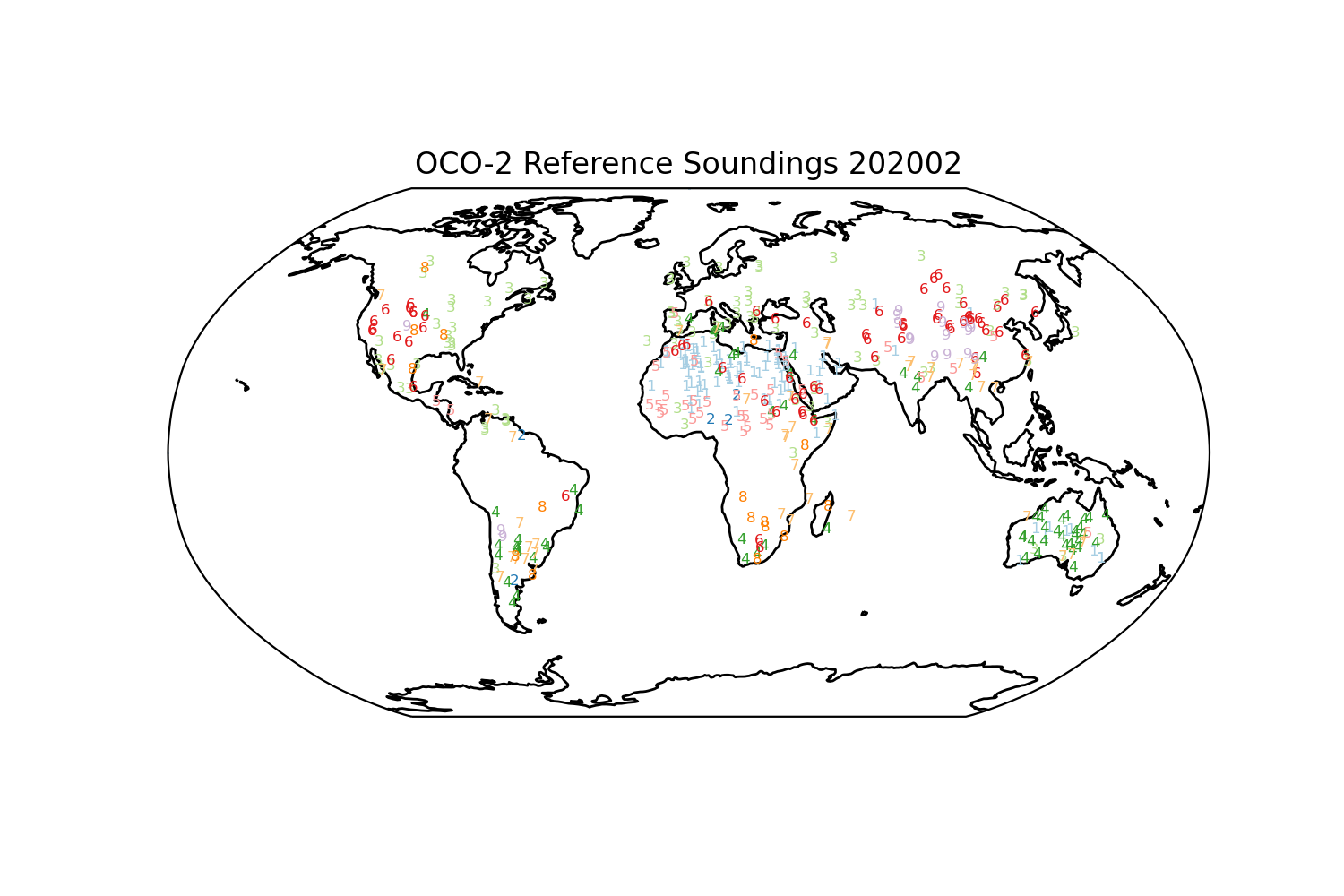}
    \caption{Locations of reference soundings for OCO-2 simulation experiments during February 2020. Colors and labels indicate the geophysical template (cluster) associated with each reference sounding.}
    \label{fig:soundings}
\end{figure}

Across the clusters, there are a total of 451 unique ``reference soundings,'' shown with their cluster membership in Figure \ref{fig:soundings}. The reference soundings are selected to span a range of instrument properties, assumed a priori states, and observing conditions within and across clusters. For each reference sounding, up to $100$ stochastic replicates are generated, giving a total dataset size of $N=44{,}982$ spectral--state pairs. The replicates vary three quantities: the state vector $\bx$, drawn from that cluster's fitted state ensemble; the model discrepancy $\bfdelta$, drawn from that cluster's discrepancy distribution; and the instrument noise $\bfepsilon$. The replicates constitute the Monte Carlo sample of the simulation experiment; they do not define the discrepancy, whose distribution is estimated beforehand and independently of this sampling, as described in Section \ref{subsec:sim_framework}.

\subsubsection{Spectral inputs}

The primary inputs to our model correspond to the three spectral bands measured by the OCO-2 instrument. Specifically, the model utilizes the $\mathbf{O_2}$ \textbf{A-band} ($0.76 \, \mu\text{m}$), which primarily constrains surface pressure and the presence of clouds and aerosols. This is complemented by the \textbf{Weak} $\mathbf{CO_2}$ \textbf{band} ($1.61 \, \mu\text{m}$), which provides the main constraint on the $\rm{CO}_2$ column, and the \textbf{Strong} $\mathbf{CO_2}$ \textbf{band} ($2.06 \, \mu\text{m}$), which introduces additional sensitivity to aerosols and water vapor, in addition to CO$_2$ absorption. Let $\by = (\by^{(1)}, \by^{(2)}, \by^{(3)})$ denote the O$_2$ A-band, weak CO$_2$ band and strong CO$_2$ band respectively. For each spectral band $j \in \{1, 2, 3\}$, the input is represented as a two-channel one-dimensional signal $\by^{(j)} \in \mathbb{R}^{2 \times L_j}$, whose first channel contains wavelengths and whose second channel contains normalized radiances.

The $2298$ spectral channels are $835$, $652$ and $811$ channels in the O$_2$ A-band, weak CO$_2$ and strong CO$_2$ bands respectively. Each band is divided by its own maximum radiance $m$, and the pair $(\lfloor\log_{10} m\rfloor,\; m\,10^{-1-\lfloor\log_{10} m\rfloor})$ (order of magnitude and the mantissa) is retained as two constant features, so that the absolute radiance scale is exactly recoverable by the network. Each band is then zero-padded on the right to a common length of $1024$. Table \ref{tab:inputs_spectral} summarizes the data and the preprocessing applied.

\begin{table}[htbp]
\centering
\begin{tabular}{lrrl}
\toprule
Band & Channels kept & Wavelengths & Fed to the network as \\
\midrule
$\mathrm{O_2}$ A-band
& 835
& 0.7592--0.7714 $\mu$m
& Wavelength + normalized radiance, padding \\

Weak $\mathrm{CO_2}$
& 652
& 1.5980--1.6178 $\mu$m
& As above \\

Strong $\mathrm{CO_2}$
& 811
& 2.0477--2.0798 $\mu$m
& As above \\
\bottomrule
\end{tabular}
 \caption{Summary of spectral inputs with transformations applied before passing them to the networks. }
 \label{tab:inputs_spectral}
\end{table}

\subsubsection{Auxiliary inputs}

Beside the spectra the model is conditioned on a small set of scalar auxiliary variables, listed in Table \ref{tab:inputs_aux}. Every one of them is known before the retrieval is run: the observing geometry (solar and viewing zenith angle, and the air mass on which the optical path length depends), the footprint index identifying which cross-track element made the observation, the meteorological prior surface pressure, and the a priori column amount and AODs that the operational retrieval is itself given. Aerosol composition enters as two categorical variables naming the two aerosol categories the retrieval was configured with for the scene, drawn from dust, sulfate, sea salt, organic carbon; the corresponding optical depths enter as two continuous prior variables. All of these are quantities typically available to ACOS Level 2 Full Physics (L2FP). Continuous variables are standardized using means and standard deviations computed from the training soundings, and categorical variables are mapped to learned embedding tables. Let $\ba$ be a vector containing these auxiliary features, so that the full input to our model is

\begin{equation}\label{eq:training}
  \bu_i = (\by^{(1)}_i, \by^{(2)}_i, \by^{(3)}_i, \ba_i), \quad \{\bu_i\}_{i=1}^{N}
\end{equation}

\begin{table}[htbp]
\centering
\small
\setlength{\tabcolsep}{4pt}
\begin{tabular}{llll}
\toprule
Variable & Unit / levels & Range (training) & Source \\
\midrule
Solar zenith angle
& degrees
& 21.95 to 72.93
& Level 1b (L1b) geolocation \\

Solar air mass $\sec\theta_0$
& dimensionless
& 1.078 to 3.407
& Derived from geometry \\

Viewing zenith angle
& degrees
& 0.253 to 1.201
& Derived from geometry \\

Viewing air mass $\sec\theta$
& dimensionless
& 1.00001 to 1.00022
& Derived from geometry \\

Total air mass
& dimensionless
& 2.078 to 4.408
& L1b, per sounding \\

Prior surface pressure
& Pa
& 55,422 to 102,363
& Meteorological prior \\

Prior 	XCO$_2$
& ppm
& 407.60 to 415.36
& ACOS a priori state \\

Prior AOD 1
& log optical depth
& $-5.28$ to $0.36$
& ACOS a priori state \\

Prior AOD 2
& log optical depth
& $-5.86$ to $-1.85$
& ACOS a priori state \\

Footprint
& 8 levels
&
& L1b instrument \\

Aerosol type 1, 2
& 4 levels
&
& ACOS configuration \\
\bottomrule
\end{tabular}
\caption{\textbf{Summary of auxiliary inputs passed to the networks.} Every variable is
available before the full-physics retrieval is run. Continuous variables are standardized with means and standard deviations computed from the
training soundings alone; categorical variables are mapped to learned embeddings. Ranges are over the training partition.}
\label{tab:inputs_aux}
\end{table}

\subsection{Preprocessing and splits}
\label{subsec:preprocessing}

Prior to training, the high-dimensional spectral data are preprocessed to stabilize gradient descent.

We consider two partitions of the $44{,}982$ soundings, which answer different questions.

The first partitions at the level of the individual replicate: the test set is the one delivered with the simulation ensemble, validation is drawn from the remaining soundings, and all $451$ reference soundings therefore contribute replicates to every partition. Results computed on it measure how well an amortized retrieval interpolates among perturbations of a scene it has already seen: a meaningful quantity, since it is what an operational system does when it reprocesses a scene under a revised state estimate.

The second partitions at the level of the reference sounding, so that no scene appears in more than one set. It is described in Section \ref{subsec:group_split}, where the results computed on it are reported. Each result in this paper indicates which partition was used.

The spectral intensities in each band are scaled by their own maximum, with two constants retained to recover the absolute scale; this is a per-sounding operation. The target and the continuous state-vector elements are standardized and aerosol types are integer-encoded, with means, standard deviations and category vocabularies estimated from the training partition alone.

\section{Methodology: Amortized probabilistic inference}
\label{sec:method}

Standard operational retrieval algorithms seek the state vector $\hat{\bx}$ that minimizes
\begin{equation}\label{eq:cost_retrival}
    \chi^2(\bx) = (\by - \bF(\bx))^\top \bS_{\epsilon}^{-1} (\by - \bF(\bx)) + (\bx-\bx_a)^\top \bS_a^{-1}(\bx - \bx_a),
\end{equation}
where $\bx_a$ and $\bS_a$ are the prior mean and covariance, and $\bS_\epsilon$ is the measurement noise covariance. This cost function is minimized for each individual measurement via iterative optimization. The process is computationally expensive, scaling linearly with the number of soundings. In contrast, our approach relies on \textit{amortized inference}. By investing computational resources upfront during the training of a neural network, we learn a direct mapping from the spectral space to the state space. Once trained, inference becomes a computationally trivial forward pass, decoupling the processing time from the complexity of the radiative transfer physics.

In this section, we detail the neural architecture used to approximate this mapping and describe two methods for elevating the network's output from a simple point estimate to a fully probabilistic posterior distribution.

\subsection{Network architecture}
\label{sec:network}

\begin{figure}[htpb]
    \centering
    \includegraphics[width=0.95\linewidth]{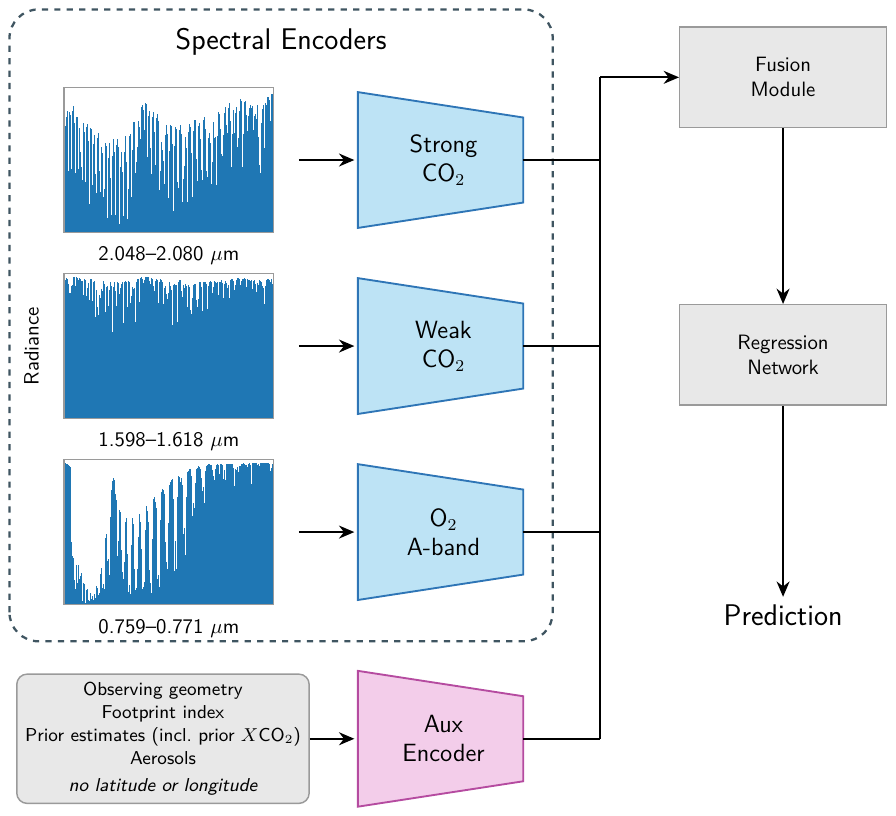}
    \caption{\textbf{Network architecture.} Each of the three OCO-2 spectral bands is processed by its own 1D convolutional encoder. The twelve auxiliary variables of Table \ref{tab:inputs_aux} (observing geometry, footprint index, the meteorological and retrieval priors, and aerosol type and loading) are embedded by a separate auxiliary encoder. No geolocation or date is supplied. The resulting four embeddings are fused through a Transformer-based attention block and passed to a task-specific Multi-layer Perceptron (MLP) head for scalar or profile prediction.}
    \label{fig:net}
\end{figure}

To capture the complex nonlinear dependencies between the spectral bands and the atmospheric state, we employ a multi-branch deep neural network. A summary of our model is shown in Figure~\ref{fig:net}. The architecture consists of three dedicated spectral encoders and one auxiliary encoder, which fuse into a shared representation before generating the final prediction.

Each spectral branch is processed by the same convolutional encoder architecture, with independent parameters across bands, which is a common choice for spectroscopic data \citep{fabbro2018,malmgren-hansen2019}. The flattened convolutional features are concatenated with the two normalizing constants, and passed through a multilayer perceptron (MLP) to produce a band embedding, $\bh^{(j)} \in \mathbb{R}^E$.

The auxiliary encoder processes the variables described in Table \ref{tab:inputs_aux}, where the module produces per-variable embeddings for the categorical features and a linear projection of continuous variables \citep{gorishniy2021,guo2016entity}. The resulting feature vector is $\bh^{(\text{aux})} \in \mathbb{R}^E$.

The band embeddings and the auxiliary embeddings are stacked into a sequence of tokens,
\begin{equation}\label{eq:features}
    \bH = (\bh^{(1)}, \bh^{(2)}, \bh^{(3)}, \bh^{\rm{(aux)}}) \in \mathbb{R}^{4 \times E}.
\end{equation}

To allow interactions among the spectral bands and auxiliary representations, we fuse the tokens using a Transformer-style attention module consisting of layer normalization followed by $L$ Transformer encoder layers with multi-head self-attention. The fused output is flattened to form a single latent representation in $\mathbb{R}^{4E}$. We refer to this stage as the fusion module (Figure~\ref{fig:net}). It is the only point at which the four branches interact: the three spectral encoders and the auxiliary encoder have independent parameters and no branch observes another's input, so the attention \citep{vaswani2017attention,nagrani2021attention} here is what allows the model to weight the bands against one another conditionally on the observing conditions carried by the auxiliary token.

Finally, the representation is passed through the regression network of Figure~\ref{fig:net}: a task-specific prediction head, an MLP \citep{bengio2013representation}, which has $D = 1$ outputs for scalar XCO$_2$ prediction, $D=20$ for full-profile prediction, and $D=4$ for last-4-layers prediction. It is the only component that differs across the three tasks. Full layer-by-layer specification for the encoders, the fusion module and the prediction head is given in~\ref{app:implementation}. More specifically, given training data $\{ (\bu_i, \bx_i) \}_{i=1}^{N_{\text{train}}}$ defined in Equation \ref{eq:training} and Section \ref{subsec:preprocessing}, where $\bx_i \in \mathbb{R}^D$ is the target state, the deterministic network defines a map $f_{\bftheta}(\bu) \in \mathbb{R}^D $, and we train this backbone as a point estimator by minimizing the empirical mean squared error:
\begin{equation}
    \hat{\theta} = \arg\min_{\theta} \frac{1}{n_{\text{train}}} \sum_{i=1}^{n_{\text{train}}} ||\bx_i - f_{\bftheta}(\bu_i)||^2_2
    \label{eq:mseloss}
\end{equation}

In all experiments, model hyperparameters such as embedding dimension, number of attention heads, number of fusion layers, learning rate and weight decay were selected using Bayesian Optimization on a validation set. Models were trained from the ground-up for each of the tasks: XCO$_2$, full $20D$ CO$_2$, and near-surface profile retrieval.

\subsubsection{Motivation for multi-branch design}

The decision to encode the spectral bands separately is motivated by the distinct physical processes governing each window of the electromagnetic spectrum. Specifically, the $\mathbf{O_2}$ \textbf{A-band} ($\sim$0.76 µm) captures strong oxygen absorption lines, providing critical constraints on the photon path length, cloud coverage, and surface pressure. When combined with solar zenith angle data, this band effectively anchors the geometric light path. Meanwhile, the \textbf{Weak} $\mathbf{CO_2}$ \textbf{band} ($\sim$1.61 µm) is sensitive primarily to the column-averaged dry air mole fraction of $\rm{CO}_2$. Because it experiences minimal interference from water vapor, it offers a relatively clean signal of the target variable. Finally, the \textbf{Strong} $\mathbf{CO_2}$ \textbf{band} ($\sim$2.06 µm), while also rich in $\rm{CO}_2$ information, is heavily influenced by water vapor and aerosol scattering. By isolating this input into its own branch, the network is forced to learn specific features related to aerosol interference without confounding the cleaner $\rm{CO}_2$ signal derived from the weak band.

\subsection{Uncertainty quantification approaches}
\label{sec:methods_UQ}

A standard neural network trained with mean squared error (MSE) loss \eqref{eq:mseloss} provides a point estimate for predictions but fails to quantify the confidence of that prediction. Given the multiple downstream uses for CO$_2$ retrievals outlined in Section \ref{sec:intro} we extend our architecture using two distinct UQ methods: the post-hoc Laplace approximation and normalizing flows.

\subsubsection{Method A: Laplace approximation}

After training the deterministic network to convergence, we place a Laplace approximation on the final linear layer while keeping the convolutional and attention-based feature extractor fixed. Let $\bfbeta$ denote the parameters of the final layer. The Laplace approximation replaces the exact posterior over $\bfbeta$ by a Gaussian centered around its maximum a posteriori (MAP) estimate,
\begin{equation*}
    q(\bfbeta) = \normal(\bfbeta; \hat{\bfbeta}_{\mathrm{MAP}}, \bfSigma_{\bfbeta})
\end{equation*}
where the mean $\hat{\bfbeta}_{\mathrm{MAP}}$ is given by the fitted value after deterministic training, and the covariance matrix $\bfSigma_{\beta}$ is obtained from a local second-order approximation to the loss around $\hat{\bfbeta}_{\mathrm{MAP}}$. The role of $q(\bfbeta)$ is to convert the deterministic point estimate into a local distribution over last-layer parameters, placing a normal posterior centered at the trained weights, and with covariance matrix $\bfSigma_{\bfbeta}$ encoding the curvature of the loss function: directions of high curvature correspond to weights that are tightly constrained by the data and hence have small posterior variance, while flatter directions correspond to greater posterior spread.

This posterior covariance over the last-layer weights induces the \emph{epistemic} component of predictive uncertainty, namely uncertainty due to limited knowledge of the model parameters \citep{kendall2017uncertainties,laplace2021}. In contrast, \emph{aleatoric} uncertainty represents irreducible variability in the observations, such as measurement noise or unresolved physical variability, and must be modeled separately \citep{hullermeier2021aleatoric}.

For a new input $\bu_*$, the network output is a $D$-dimensional profile
\begin{equation*}
    f_{\beta}(\bu_*) \in \mathbb{R}^{D},
\end{equation*}
and we propagate uncertainty in $\bfbeta$ to output space by Monte Carlo sampling from the Laplace posterior. Specifically, if
\begin{equation*}
    \bfbeta^{(s)} \sim q(\bfbeta), \qquad s=1,\dots,S,
\end{equation*}
where $s$ indexes Monte Carlo draws. The corresponding sampled profile predictions are
\begin{equation*}
    \hat{\bx}_*^{(s)} = f_{\bfbeta^{(s)}}(\bu_*) \in \mathbb{R}^{D}.
\end{equation*}

The predictive mean is approximated by
\begin{equation*}
    \hat{\bfmu}_*(\bu_*) =\frac{1}{S}\sum_{s=1}^S \hat{\bx}_*^{(s)},
\end{equation*}
and the epistemic covariance is estimated by the sample covariance
\begin{equation*}
\widehat{\bfSigma}_{\mathrm{epi}}(\mathbf{u}_*)= \frac{1}{S-1}\sum_{s=1}^S \left(\hat{\bx}_*^{(s)}-\hat{\bfmu}_*(\bu_*)\right) \left(\hat{\mathbf{x}}_*^{(s)}-\hat{\bfmu}_*(\bu_*)\right)^\top.
\end{equation*}

Since the Laplace approximation over the final layer captures only epistemic uncertainty, we estimate the aleatoric component from validation residuals. Let $\{(\bu_i,\bx_i)\}_{i=1}^{n_{\mathrm{val}}}$ denote a validation set, and let $\hat \bfmu_i$ and $\widehat{\bfSigma}_{\mathrm{epi}}(\bu_i)$ be the corresponding Laplace predictive mean and epistemic covariance. We estimate the aleatoric covariance diagonally, coordinate-wise, via
\begin{equation*}
    \hat{\sigma}^2_{\mathrm{aleat},j} =   \frac{1}{n_{\rm val}}\sum_{i=1}^{n_{\rm val}}\left( (x_{ij}-\hat\mu_{ij})^2 - [\hat\bfSigma_{\rm epi}(\bu_i)]_{jj}\right).
\end{equation*}
for $j = 1, \dots, D$, so that we define $\widehat{\bfSigma}_{\mathrm{aleat}} = \rm{diag}(\hat{\sigma}^2_{\mathrm{aleat},1},\dots,\hat{\sigma}^2_{\mathrm{aleat},D})$.  The difference can in principle be negative, if the epistemic variance alone exceeds the residual variance, so each $\hat{\sigma}^2_{\mathrm{aleat},j}$ is floored at $10^{-10}$ in standardized units. The floor is not active for any model reported here: the last-layer epistemic variance is small compared with the residual variance, and the estimated aleatoric variances range from $0.23$ to $0.77$ in standardized units. The final predictive covariance is therefore

\begin{equation*}
\widehat{\bfSigma}_{\mathrm{pred}}(\bu_*) = \widehat{\bfSigma}_{\mathrm{epi}}(\bu_*) +
\widehat{\bfSigma}_{\mathrm{aleat}}.
\end{equation*}

Thus, the total predictive uncertainty is decomposed into an epistemic term arising from uncertainty in the last-layer parameters and an aleatoric term capturing residual variability not explained by parameter uncertainty alone.

This approach is readily adaptable for $D=1$ for the scalar XCO$_2$ task. We train the network to predict the scalar quantity, and consider scalar functions/distributions as appropriate above.

\subsubsection{Method B: Normalizing flows}

To address the limitations of Gaussian assumptions, which is a key critique of operational retrievals, we employ normalizing flows. Normalizing flows learn a series of invertible, differentiable transformations $f$ that map a simple base distribution (e.g., a standard spherical Gaussian) to a complex, potentially multi-modal, target distribution.

Conditioned on the spectral embeddings $\mathbf{h}(\mathbf{y})$ produced by our network, the flow defines the posterior density as:
\begin{equation}
    p(\mathbf{x} | \mathbf{y}) = p_0(z) \left| \det \frac{\partial f^{-1}(\mathbf{x}; \mathbf{h}(\mathbf{y}))}{\partial \mathbf{x}} \right|
\end{equation}
where $z = f^{-1}(\mathbf{x})$ is the variable in the latent Gaussian space. By chaining multiple non-linear bijectors (such as masked autoregressive flows or affine coupling layers), the model can represent highly skewed, asymmetric, or multi-modal posteriors, which traditional Gaussian approximations fail to represent.

For our implementation, we retain the deterministic backbone as a context network, producing embeddings in $\mathbb{R}^{4E}$, and replace the MLP prediction head by a conditional flow on $\mathbb{R}^D$ (where $D = 1, 4, 20$ depending on the task). For the profile tasks ($D = 4, 20$), the fused feature vector produced by the backbone serves as the conditioning context for a stack of piecewise-rational-quadratic coupling transforms implemented with the \verb|nflows| package \citep{nflows2020}, with alternating binary masks and random permutations between layers so that all profile coordinates are transformed. For scalar XCO$_2$ ($D = 1$) a coupling transform has no coordinates to split, and the flow is instead a one-dimensional autoregressive rational-quadratic spline conditioned on the flattened encoder output without the fusion module (Appendix~\ref{app:implementation}).

\section{Numerical comparisons}
\label{sec:experiments}

See~\ref{sec:metrics} for a brief discussion of the metrics used to evaluate the models. All evaluations are performed on held-out simulated soundings.

\subsection{Operational retrieval (L2 Full Physics)}
\label{subsec:baselines}

The primary baseline is the ACOS L2FP retrieval, version 10 \citep{ocoatbd_v10, odell2018acos, taylor2023evaluating}, the same build used to generate the simulation ensemble; the spectroscopy is the absorption coefficient table set (ABSCO v5.1) adopted for the version-10 Level-2 algorithm \citep{payne2020}. Every retrieval reported here was performed on simulated radiances. The reported XCO$_2$ is the raw state estimate. The empirical bias correction applied to the delivered operational product \citep{v10dug} is fitted against coincidences with ground-based validation and cannot be transferred to a simulation whose reference state is synthetic, so it is not applied. The comparison is therefore between two estimators of the state given the radiances, and we refer to the comparator as ACOS L2FP v10 (raw, on simulated radiances), abbreviated ACOS L2FP. The retrieval was run on each of our $44{,}982$ simulated spectra together with its meteorology.

We apply no quality filtering and report every sounding. The simulation ensemble quality flag, which was not used in the simulation dataset construction and is based primarily on goodness of spectral fit, is good for 89.8\% of the soundings, and restricting to those soundings changes the comparator's XCO$_2$ root-mean-square error (RMSE)  from $2.91$ to $2.73$ ppm, which alters no comparison in the February partitions. Non-convergence is not a factor: the retrieval converges for $99.9\%$ of soundings, with a median of four iterations. The ACOS L2FP prior has a mean $\mathbf{x}_a$ that varies by sounding and a covariance $\mathbf{S}_a$ that does not; we score $\mathcal{N}(\mathbf{x}_a, \mathbf{S}_a)$ as a baseline retrieval for every target (Tables~\ref{tab:splitcompare} and~\ref{tab:months}). Its prior standard deviation grows from $1.4$~ppm at the top level to $48$~ppm at the surface, so it covers the truth near the surface almost always while carrying little information there.

The minimization is performed using an iterative Levenberg--Marquardt optimizer, requiring repeated executions of the computationally expensive forward model $\mathbf{F}$. Furthermore, the posterior uncertainty $\hat{\mathbf{S}}$ is computed as the inverse Hessian of the cost function at the solution. Crucially, this approach assumes that the forward model is perfect ($\boldsymbol{\delta} = \bfzero$) and that the posterior distribution is locally Gaussian.

That assumption has a direct consequence for the uncertainties we compare against. An OE posterior computed under a perfect forward model contains no term representing forward-model error, while every sounding in this ensemble carries a non-zero discrepancy $\bfdelta$. The comparator is therefore being asked to quantify an error source its own covariance does not represent by construction. The under-coverage reported in Sections \ref{sec:oco2_prediction}--\ref{sec:last4} should be read in that light: it is a structural property of applying a perfect-forward-model posterior to data containing model discrepancy.

\subsection{Predicting XCO$_2$}\label{sec:oco2_prediction}

Table \ref{tab:xco2_prediction} compares the ACOS L2FP (v10) retrieval, the Laplace approximation built on top of the deterministic regressor, and the conditional normalizing flow for scalar XCO$_2$ prediction. The normalizing flow attains the lowest RMSE and mean absolute error (MAE), with the Laplace approximation a close second; the difference between the two is not resolved by these data (paired difference is $0.017$ ppm while standard error is $0.011$). This is notable because the flow is not trained with an MSE objective but by maximizing the conditional likelihood: it remains competitive in point prediction even though it is optimized for probabilistic performance. As expected the flow also attains the best negative log-likelihood (NLL). ACOS L2FP performs substantially worse than the other two in the reported metrics.

\begin{table*}[htpb]
\centering
\setlength{\tabcolsep}{8pt}
\renewcommand{\arraystretch}{1.1}
\begin{tabular}{lcccc}
\toprule
Method
& RMSE [ppm]
& MAE [ppm]
& NLL
& 95\% coverage \\
\midrule
ACOS L2FP
& 2.91 (0.05)
& 2.32 (0.04)
& 21.50 (0.81)
& 21.9 \\

Laplace
& 1.48 (0.03)
& 1.12 (0.02)
& 0.64 (0.02)
& 94.7  \\

Normalizing flow
& \textbf{1.46} (0.03)
& \textbf{1.08} (0.02)
& \textbf{0.52} (0.02)
& 94.8 \\
\bottomrule
\end{tabular}
\caption{\textbf{XCO$_2$ retrieval, within-reference test partition}. RMSE and MAE in ppm at each method's own optimal point estimate; NLL per sounding; coverage in percent at nominal 95\%. Parentheses are bootstrap standard errors resampling reference soundings. Definitions in ~\ref{sec:metrics}.}
\label{tab:xco2_prediction}
\end{table*}

\begin{figure}[htpb]
    \centering
    \includegraphics[width=\linewidth]{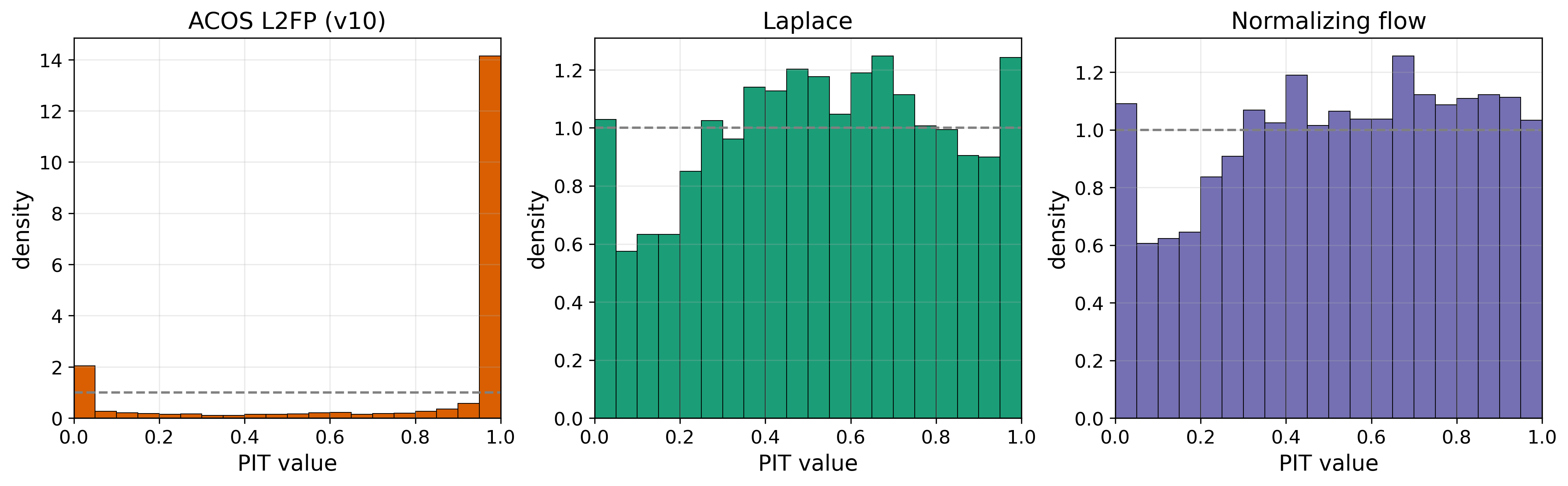}
    \caption{
    PIT histograms for scalar XCO$_{2}$ prediction across the three methods. A perfectly calibrated predictive distribution produces a uniform histogram (dashed line). ACOS L2FP (left) shows severe concentration of mass near 0 and 1, indicating that the true value frequently falls outside the predicted intervals. The Laplace approximation (center) shows a hump-shaped histogram peaking near 0.5. The normalizing flow (right) produces the histogram closest to uniform, with only mild deviations.}
    \label{fig:pit_xco2}
\end{figure}

\begin{figure}[htpb]
    \centering
    \includegraphics[width=0.5\linewidth]{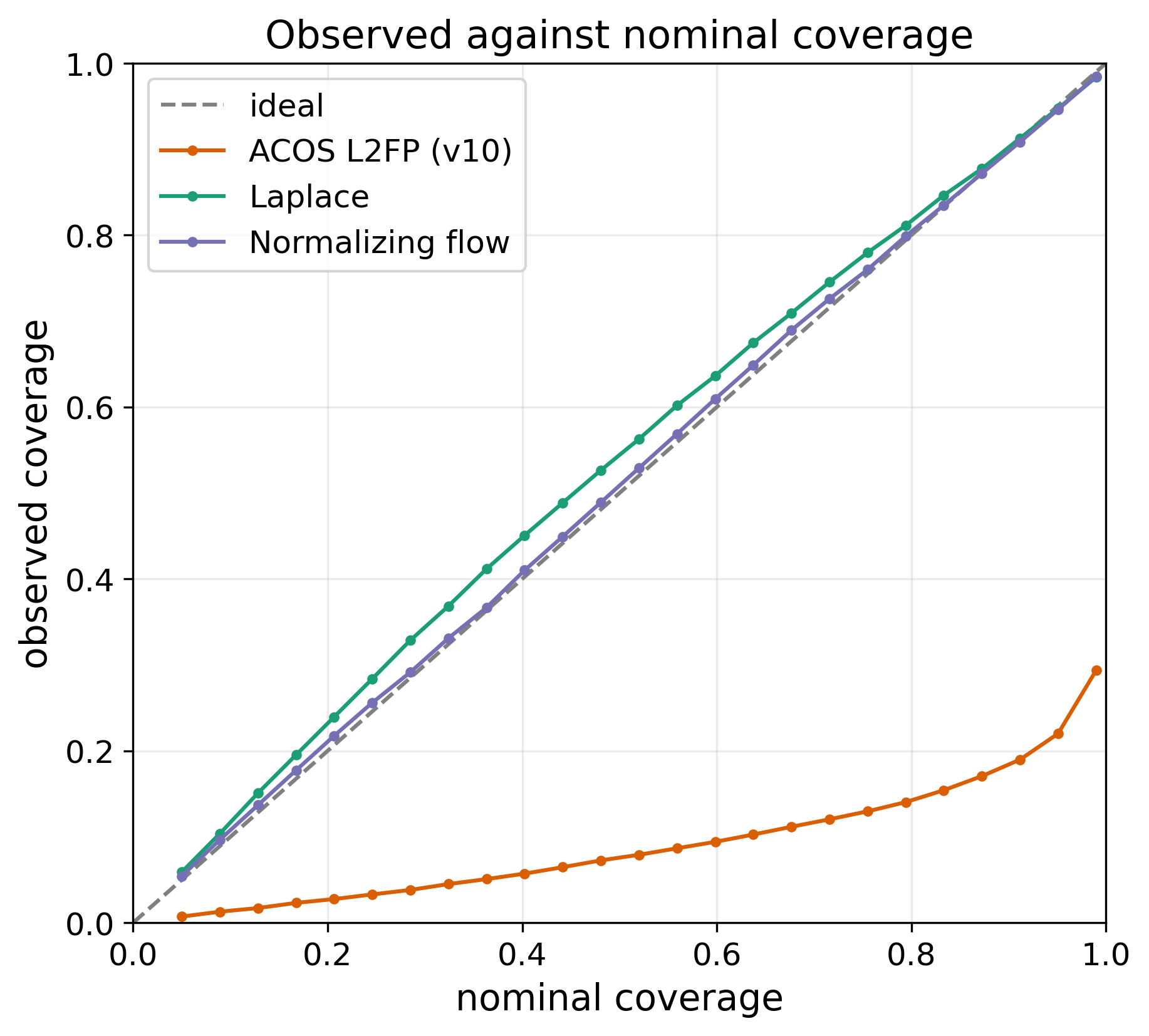}
    \caption{
    Empirical coverage curves for scalar XCO$_{2}$ prediction. A perfectly calibrated method follows the diagonal (dashed line). ACOS L2FP retrieval lies far below the diagonal across all nominal levels, confirming severe undercoverage consistent with its non-uniform PIT histogram. The Laplace approximation lies above the diagonal at low and moderate nominal levels, reflecting conservative over-coverage. The normalizing flow tracks the diagonal most closely among all three methods.}
    \label{fig:coverage_xco2}
\end{figure}

Calibration diagnostics (Figures \ref{fig:pit_xco2}--\ref{fig:coverage_xco2}) reinforce this picture. The probability integral transform (PIT) histogram for ACOS L2FP is strongly non-uniform, with most mass concentrated near $1$ and a smaller excess near $0$. A uniform PIT histogram is expected under a calibrated predictive distribution. The excess near the boundaries suggest that the retrieval distribution might be too narrow, while asymmetry towards $1$ points to systematic under-prediction. This is consistent with the empirical coverage values observed in Table \ref{tab:xco2_prediction} and the consistent low observed coverage values in Figure \ref{fig:coverage_xco2}.

The Laplace approximation is much better calibrated than ACOS L2FP, but its PIT histogram remains mildly hump shaped, with a peak around $0.5$. This indicates over-dispersion: the predictive distributions are somewhat too wide. The coverage curve (Figure \ref{fig:coverage_xco2}) supports this conclusion, as the observed coverage lies above the nominal diagonal. What remains is a mild excess at moderate levels rather than a mis-specified scale. Its gap to the flow is then not one of scale but of shape, see Section \ref{sec:nongauss}.

Among the three methods, the normalizing flow shows the best probabilistic calibration. Its PIT histogram is closest to uniform (see Figure \ref{fig:pit_xco2}), with mild residual deviations. These results suggest that the flow offers the best balance of accuracy and calibration: it attains the lowest RMSE, MAE and NLL of the three.

\subsection{Predicting the full CO$_2$ profile}

\begin{figure}[htpb]
    \centering
    \includegraphics[width=\linewidth]{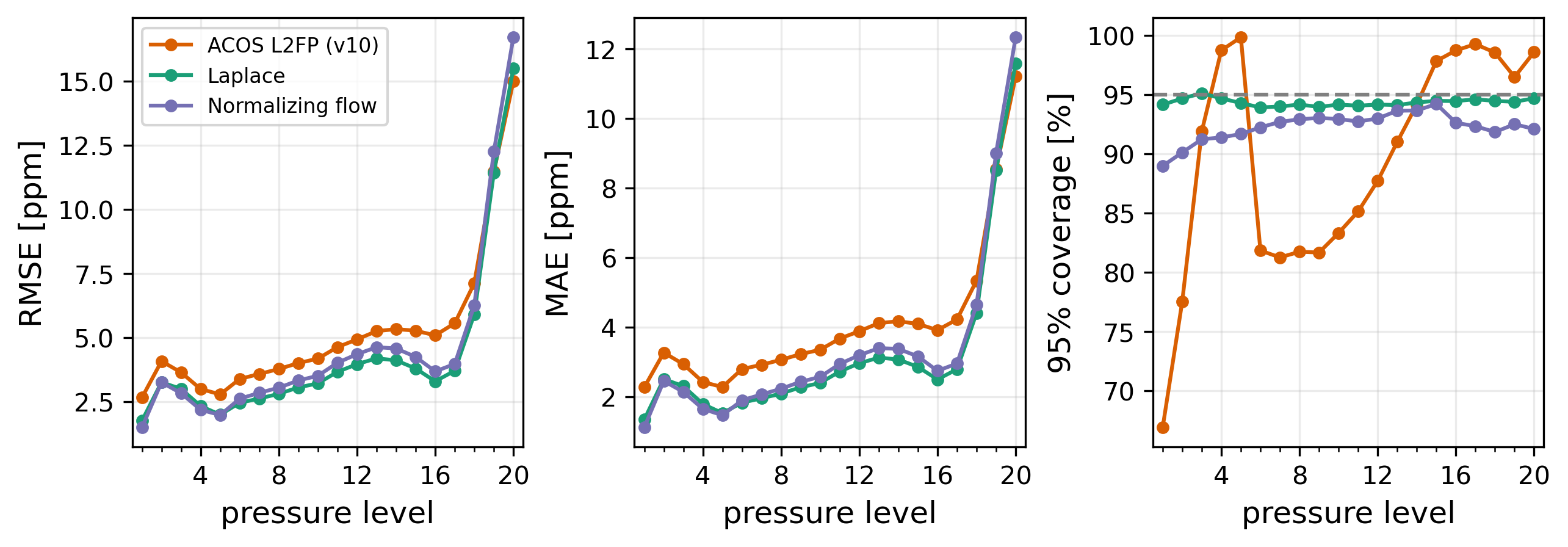}
    \caption{\textbf{Per-level performance on the 20-levels profiles,} within reference partition. RMSE (left) and MAE (center) in ppm, empirical $95\%$ marginal coverage (right), against pressure level, level 20 being nearest to surface; dashed line is nominal. MAE is at each method's own MAE-optimal point estimate: the mean for the two Gaussians, the predictive median for the flow. The learned models lead over most of the column, except for the two levels nearest the surface.  }
    \label{fig:20d_main}
\end{figure}

We next compare the operational retrieval, Laplace approximation, and the conditional normalizing flow for full $20$-dimensional CO$_2$ profile prediction. Table \ref{tab:model-comparison20D} summarizes and Figure \ref{fig:20d_main} reports per-level RMSEs, MAEs, and empirical coverages. The normalizing flow was able to capture the joint structure of uncertainty across pressure levels, achieving by far the lowest joint NLL. This is the regime in which a flexible, possibly non-Gaussian conditional density should be expected to help: even when point predictions are similar, the conditional normalizing flow model has the ability to represent non-elliptical dependence and more adaptive uncertainty across levels.

In terms of point prediction, the Laplace approximation remains competitive and typically attains the best RMSE, although the gap with the normalizing flow is small. The main difference between these two approaches is joint fidelity.

The normalizing flow tends to slightly undercover at the marginal level, with observed coverage somewhat below nominal across layers. The flow is trained to maximize joint conditional likelihood rather than to calibrate marginal intervals at each pressure level, so this is not surprising. The Laplace approximation shows mild over-coverage at moderate nominal levels, while maintaining close to nominal coverage at the $95\%$ level (Figure \ref{fig:coverage_xco2}).

An interesting exception appears near the surface. At the last two pressure levels, the ACOS L2FP can match or even outperform the learned models in RMSE and MAE. This suggests that the full-profile task may not be the right objective when interest is concentrated on a small subset of levels, and motivates the dedicated near-surface models of Section \ref{sec:last4}.

Overall, the results illustrate that the normalizing flow delivers the best representation for the full joint posterior over profiles. Downstream applications that depend on coherent uncertainty across multiple pressure levels would then gain a practical advantage while using this model.  If interest is in a specific summary of the full-profile (e.g., XCO$_2$ or only one or a few pressure levels), then one would train the normalizing flow to predict exactly that summary.

\begin{table*}[htpb]
\centering
\setlength{\tabcolsep}{8pt}
\renewcommand{\arraystretch}{1.1}
\begin{tabular}{lcccc}
\toprule
Method
& RMSE [ppm]
& MAE [ppm]
& NLL
& cov. 95\% (joint/marg) \\
\midrule
ACOS L2FP
& 6.00 (0.09)
& 4.08 (0.05)
& 45.19 (1.02)
& 42.4 / 89.6 \\

Laplace
& \textbf{5.39} (0.10)
& \textbf{3.22} (0.04)
& 21.51 (0.29)
& 83.0 / 94.4\\

Normalizing flow
& 5.79 (0.11)
& 3.38 (0.05)
& $-$\textbf{46.44} (0.55)
& 96.0 / 92.3\\
\bottomrule
\end{tabular}
\caption{\textbf{20-level CO$_2$ profile, within-reference test partition.} RMSE and MAE are pooled over soundings and levels, in ppm; NLL is the joint density over the 20 levels; coverage is printed joint / marginal, the joint figure being the $\chi^2$ coverage of the full 20-dimensional predictive and the marginal one the average over levels. Parentheses are bootstrap standard errors resampling reference soundings.}
\label{tab:model-comparison20D}
\end{table*}

\subsection{Near-surface levels}\label{sec:last4}

\begin{figure}[htpb]
    \centering
    \includegraphics[width=\linewidth]{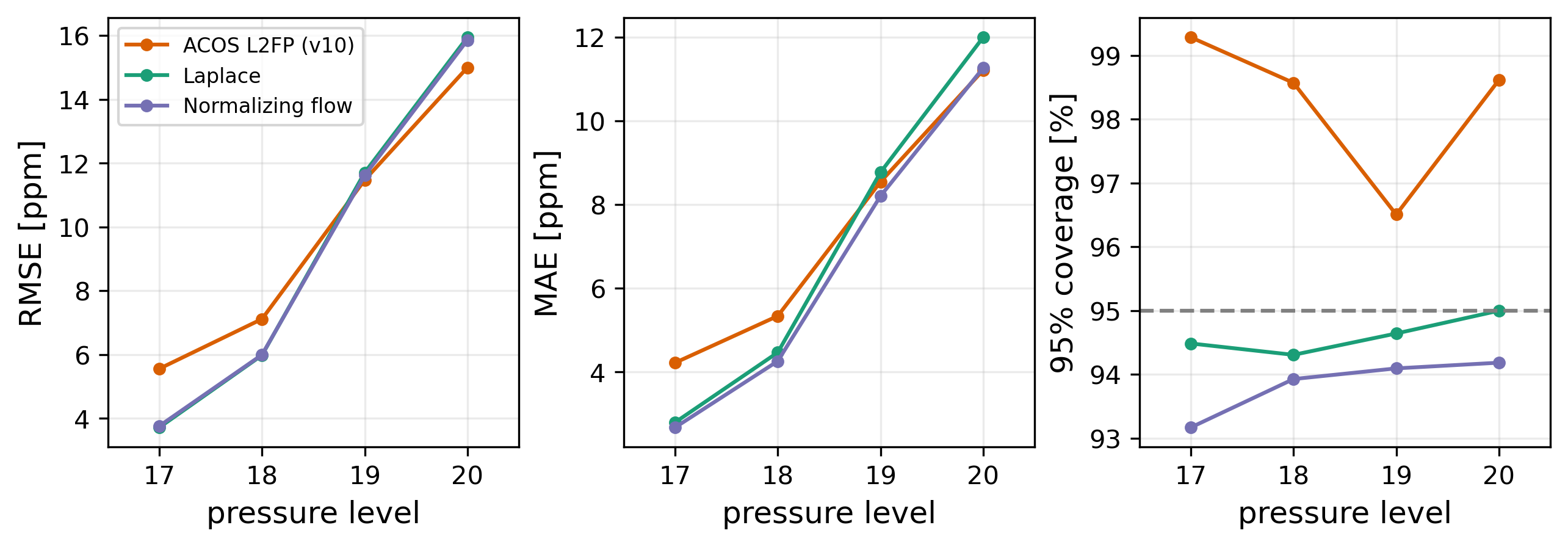}
    \caption{\textbf{Per-level performance on the four near-surface levels (17--20),} within reference partition, for models trained directly on those levels. Axes as in Figure \ref{fig:20d_main}.}
    \label{fig:last4_main}
\end{figure}

The results for the four vertical levels nearest the surface sharpen the distinction between modeling the full profile jointly and modeling a target subset of scientific interest. While the full-profile normalizing flow is the most faithful model, its induced marginals over near-surface layers are not optimal. In this setting a dedicated $4$-output normalizing flow attains the best joint NLL by a wide margin and the best pooled MAE, and leads on MAE at three of the four levels; pooled RMSE is a three-way tie, with the operational retrieval ahead at the two levels closest to the surface; see Figure \ref{fig:last4_main} and Table \ref{tab:model-comparison-last4}. Note that, for downstream applications focused on the near-surface portion of the column, it is preferable to train the probabilistic retrieval model directly on those levels.

\begin{table*}[htpb]
\centering
\setlength{\tabcolsep}{8pt}
\renewcommand{\arraystretch}{1.1}
\begin{tabular}{lcccc}
\toprule
Method
& RMSE [ppm]
& MAE [ppm]
& NLL
& cov. 95\% (joint/marg) \\
\midrule
ACOS L2FP
& 10.47 (0.20)
& 7.33 (0.12)
& 5.29 (0.05)
& 92.5 / 98.2 \\

Laplace
& 10.50 (0.21)
& 7.01 (0.12)
& 4.37 (0.07)
& 89.9 / 94.6\\

Normalizing flow
& \textbf{10.45} (0.25)
& \textbf{6.60} (0.13)
& $-$\textbf{6.29} (0.13)
& 96.9 / 93.8\\
\bottomrule
\end{tabular}
\caption{\textbf{Near-surface levels (17--20), within-reference test partition} for models trained directly on those four levels. Columns as in Table \ref{tab:model-comparison20D}, the joint coverage being over the four levels.}
\label{tab:model-comparison-last4}
\end{table*}

\subsection{Generalization to unseen reference soundings}\label{subsec:group_split}

The experiments of Sections \ref{sec:oco2_prediction}--\ref{sec:last4} partition the simulated soundings at the level of the stochastic replicate, so each of the $451$ reference soundings contributes replicates to both the training and the test set. Those results measure how well an amortized retrieval interpolates among perturbations of a scene it has already seen. We therefore repeat the principal experiments on a partition that is disjoint at the level of the reference sounding.

The $451$ reference soundings are apportioned $361 / 45 / 45$ to training, validation and test, stratified by geophysical cluster, with all replicates of a reference sounding confined to one partition; this gives $36{,}003 / 4{,}490 / 4{,}489$ soundings. Cluster 2 contains only five reference soundings and is assigned entirely to training, so the test partition spans eight of the nine clusters. All preprocessing statistics are re-estimated from the training reference soundings. Table \ref{tab:splitcompare} reports
the results.

\begin{table*}[htpb]
\centering
\small
\setlength{\tabcolsep}{5pt}
\renewcommand{\arraystretch}{1.1}
\resizebox{\columnwidth}{!}{\begin{tabular}{lrrrc@{\hskip 14pt}rrrc}
\toprule
 & \multicolumn{4}{c}{Within-reference (interpolation)} & \multicolumn{4}{c}{Reference-disjoint (generalization)} \\
\cmidrule(lr){2-5}\cmidrule(lr){6-9}
Method & RMSE & MAE & NLL & cov.\ 95 & RMSE & MAE & NLL & cov.\ 95 \\
\midrule
\multicolumn{9}{l}{\emph{Scalar XCO$_2$}} \\
ACOS L2FP (v10)  & 2.91 (0.05) & 2.32 (0.04) & 21.50 & 21.9 & 2.87 (0.12) & 2.29 (0.09) & 20.20 & 23.0 \\
Laplace          & 1.48 (0.03) & 1.12 (0.02) &  0.64 & 94.7 & 2.26 (0.39) & 1.45 (0.19) & 0.98 & 92.7\\
Normalizing flow & \textbf{1.46} (0.03) & \textbf{1.08} (0.02) & \textbf{0.52} & 94.8 & \textbf{2.11} (0.33) & \textbf{1.38} (0.17) & \textbf{0.74} & 93.4\\
Prior XCO$_2$ & 2.22 (0.08) & 1.62 (0.04) & 2.41 & 100.0 & 2.27 (0.27) & 1.63 (0.13) & 2.42 & 100.0 \\
\midrule
\multicolumn{9}{l}{\emph{$20$-level $\rm{CO}_2$ profile}} \\
ACOS L2FP (v10)  & 6.00 (0.09) & 4.08 (0.05) & 45.19 & 42.4 / 89.6 & \textbf{6.08} (0.22) & 4.14 (0.12) & 43.77 & 44.3 / 89.5 \\
Laplace          & \textbf{5.39} (0.10) & \textbf{3.22} (0.04) & 21.51 & 83.0 / 94.4 & 6.39 (0.47) & \textbf{3.72} (0.22) & 24.96 & 81.7 / 93.5 \\
Normalizing flow & 5.79 (0.11) & 3.38 (0.05) & $-$\textbf{46.44} & 96.0 / 92.3 & 6.54 (0.44) & 3.88 (0.20) & $-$\textbf{8.32} & 58.7 / 91.7 \\
Prior &6.65 (0.15) & 4.08 (0.06) & 20.54 & 95.8 / 96.8 & 6.79 (0.50) & 4.11 (0.20) & 20.54 & 96.3 / 96.9 \\
\midrule
\multicolumn{9}{l}{\emph{Near-surface levels ($17$--$20$)}} \\
ACOS L2FP (v10)  & 10.47 (0.20) & 7.33 (0.12) & 5.29 & 92.5 / 98.2 & \textbf{10.66} (0.49) & \textbf{7.52} (0.31) & 5.42 & 91.2 / 98.3 \\
Laplace          & 10.50 (0.21) & 7.01 (0.12) & 4.37 & 89.9 / 94.6 & 11.95 (0.90) & 7.80 (0.50) & 5.08 & 88.9 / 93.8 \\
Normalizing flow & \textbf{10.45} (0.25) & \textbf{6.60} (0.13) & $-$\textbf{6.29} & 96.9 / 93.8 & 12.48 (0.90) & 8.32 (0.49) & \textbf{0.53} & 94.3 / 94.3 \\
Prior &12.57 (0.31) & 8.60 (0.17) & 6.20 & 100.0 / 100.0 & 12.86 (1.01) & 8.70 (0.54) & 6.22 & 100.0 / 99.9 \\

\bottomrule
\end{tabular}
}
\caption{Accuracy, likelihood and calibration on the two partitions. RMSE and MAE are pooled over pressure levels and reported in ppm, with MAE computed at the posterior median (for the two Gaussian methods the median coincides with the mean); NLL is computed per sounding; coverage is in percent at the nominal $95\%$ level, printed as joint / marginal, where the joint figure is the $\chi^2$ coverage of the full predictive and the marginal figure is the average over levels (for XCO$_2$ the two estimate the same quantity). Parentheses are bootstrap standard errors resampling reference soundings; the reference-disjoint test partition contains $45$ of them against $451$ in the within-reference partition, so the two sets of standard errors are not directly comparable. Bold marks the best value in a column within a task; coverage is not bolded, since neither direction of departure from nominal is better. The prior rows score the operational a priori as a retrieval with no fitting: its mean state $\mathbf{x}_a$ as the point estimate and $\mathcal{N}(\mathbf{x}_a, \mathbf{S}_a)$ as the predictive distribution, with $\mathbf{S}_a$ the same for every sounding; for XCO$_2$ the a priori column amount and its variance $\mathbf{h}^\top \mathbf{S}_a \mathbf{h}$, with $\mathbf{h}$ the pressure-weighting function.}
\label{tab:splitcompare}
\end{table*}

\begin{table*}[htbp]
\centering
\small
\setlength{\tabcolsep}{5pt}
\renewcommand{\arraystretch}{1.1}
\begin{tabular}{lrrr@{\hskip 12pt}rrr}
\toprule
 & \multicolumn{3}{c}{XCO$_2$} & \multicolumn{3}{c}{Near-surface (levels 17--20)} \\
\cmidrule(lr){2-4} \cmidrule(lr){5-7}
Method & RMSE & NLL & cov.\ 95 & RMSE & NLL & cov.\ 95 \\
\midrule
\multicolumn{7}{l}{\emph{March 2020}, $n=44{,}985$} \\
\quad ACOS L2FP (v10)  & 3.08 (0.04) & 25.14 & 25.4 & \textbf{11.63} (0.20) & 6.01 & 89.1 / 97.3 \\
\quad Laplace          & 2.44 (0.13) & 1.33 & 87.7 & 16.29 (0.42) & 7.24 & 75.6 / 85.2 \\
\quad Normalizing flow & \textbf{2.42} (0.15) & \textbf{1.14} & 87.0 & 15.06 (0.47) & \textbf{2.02} & 82.6 / 88.7 \\
\quad Prior & 2.63 (0.20) & 2.43 & 100.0 & 15.28 (0.56) & 6.27 & 99.8 / 99.8 \\
\midrule
\multicolumn{7}{l}{\emph{July 2020}, $n=44{,}327$} \\
\quad ACOS L2FP (v10)  & 3.51 (0.14) & 36.20 & 26.3 & \textbf{14.18} (0.34) & 7.54 & 86.1 / 95.3 \\
\quad Laplace          & 2.76 (0.07) & 1.69 & 78.3 & 17.26 (0.22) & 7.86 & 68.8 / 80.5 \\
\quad Normalizing flow & \textbf{2.67} (0.08) & \textbf{1.55} & 79.2 & 16.43 (0.25) & \textbf{3.37} & 70.0 / 86.0 \\
\quad Prior & 2.73 (0.09) & 2.43 & 100.0 & 15.32 (0.23) & 6.29 & 100.0 / 100.0 \\
\bottomrule
\end{tabular}
\caption{\textbf{Temporal generalization.} XCO$_2$ and near-surface retrieval on two months of the same simulation ensemble that the network never saw. The two blocks are the all-February model applied to every sounding of the month, with no refitting of any kind. ACOS L2FP does not train and is the same retrieval throughout, so its rows measure the difficulty of the month rather than any model's ability to generalize, and the learned rows should be read against it. RMSE in ppm; NLL is the joint predictive density in standardized target coordinates, as in Table~\ref{tab:splitcompare}; parentheses are bootstrap standard errors resampling whole reference soundings. Coverage is the empirical frequency at a nominal $95\%$ level, joint~/~marginal for the four-dimensional target and a single number for the column average. Bold marks the best RMSE and the best NLL within a block and task. One July sounding carries a non-physical retrieved state in the operational retrieval and is excluded from the point-error columns of all methods. The prior rows are defined as in Table~\ref{tab:splitcompare}.}
\label{tab:months}
\end{table*}

\begin{figure}[htpb]
    \centering
    \includegraphics[width=\linewidth]{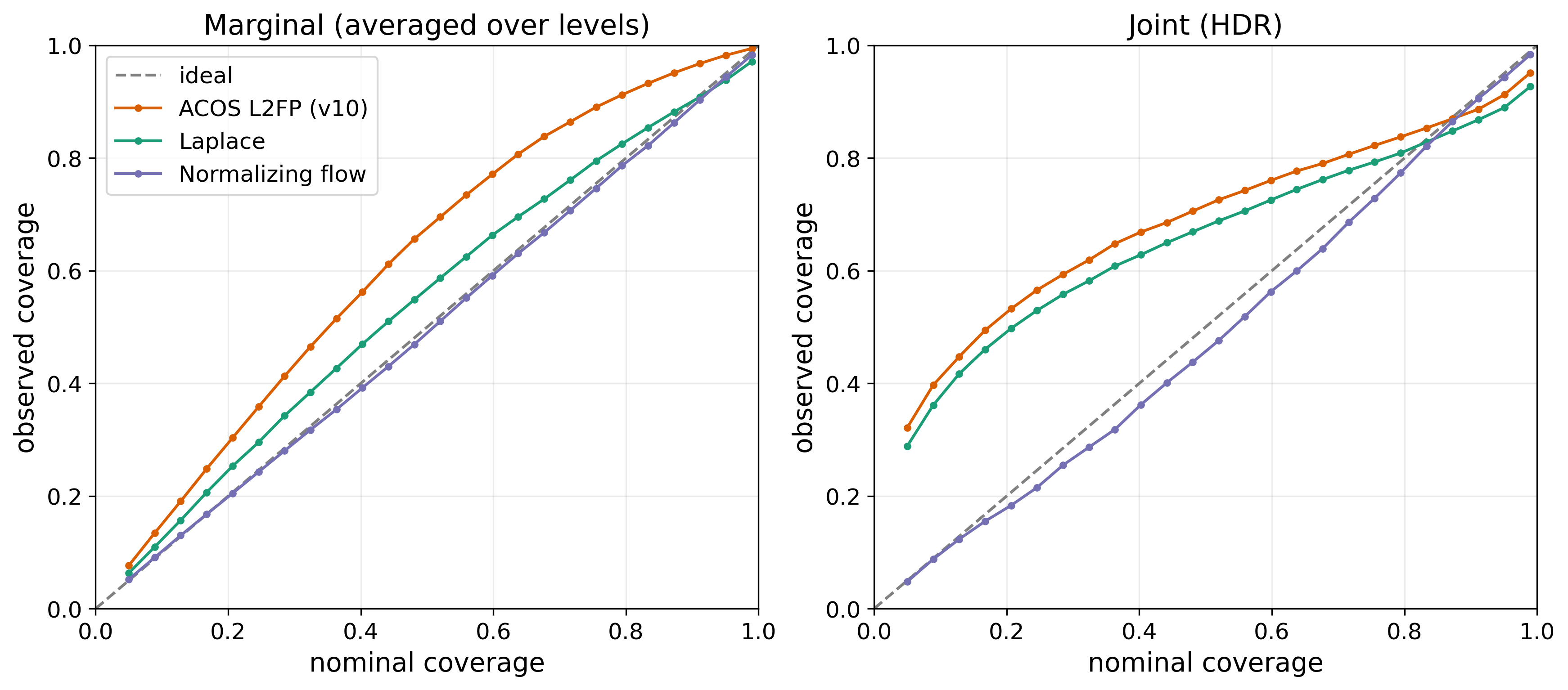}
    \caption{\textbf{Observed against nominal coverage for the four near-surface levels}, reference-disjoint test partition. Left: marginal coverage, the per-level rate averaged over the four levels. Right: joint coverage of the full four-dimensional predictive. Both are defined in \ref{sec:metrics}; the dashed diagonal is perfect calibration. The two panels separate what a per-level rate can and cannot show: ACOS L2FP over-covers marginally ($98.3\%$ at the nominal $95\%$ level) while under-covering jointly ($91.2\%$), and the Laplace approximation is near nominal marginally ($93.8\%$) and below it jointly ($88.9\%$). For the normalizing flow the two agree, at $94.3\%$. This is consistent with correct representation of between-level dependence but does not establish it.}
    \label{fig:groupsplit_coverage}
\end{figure}

On this partition, accuracy degrades while marginal calibration largely holds. For scalar XCO$_2$ the normalizing flow remains the best of the three methods on RMSE, MAE and likelihood, but its RMSE rises from $1.46$ to $2.11$ ppm, and its improvement over the prior XCO$_2$ falls from $34\%$ to $7\%$ in RMSE. On the two profile tasks the learned models no longer attain the lowest pooled RMSE: the operational retrieval does, at $6.08$ ppm against the flow's $6.54$ for the $20$-level profile and $10.66$ against $12.48$ for the near-surface levels. The likelihood ordering is unchanged on all three tasks, with the flow ahead of the Laplace approximation and both ahead of the operational retrieval. The flow's marginal coverages remain close to nominal on every task and both partitions. Its joint $95\%$ coverage on the $20$-level task is the one calibration quantity that does not transfer, falling from $96.0\%$ to $58.7\%$: the per-level dispersion carries over to unseen scenes, while the learned correlation structure across levels does not. Figure \ref{fig:groupsplit_coverage} shows this distinction as curves and shows that for the near-surface task, the normalizing flow produces the better calibrated predictive posteriors. The a priori rows show what the retrievals add. In RMSE the flow is more accurate than the prior for XCO$_2$ on both partitions and both unseen months, and the Laplace approximation within reference and in March; it is level with the prior on unseen reference soundings ($2.26$ against $2.27$~ppm) and behind it in July ($2.76$ against $2.73$~ppm). Every method is more accurate than the prior for the $20$-level profile. Near the surface the margin is smaller, and on the unseen months the learned methods lose it: in July both are less accurate than the prior ($16.43$ and $17.26$ against $15.32$~ppm), as the Laplace approximation is in March. As a predictive distribution the prior is wide but not misplaced: for the $20$-level profile its NLL ($20.54$) is below that of the Laplace approximation ($21.51$ and $24.96$) and its joint coverage is close to nominal, so only the flow improves on it as a density over the full profile.

\begin{figure}[htpb]
    \centering
    \includegraphics[width=\linewidth]{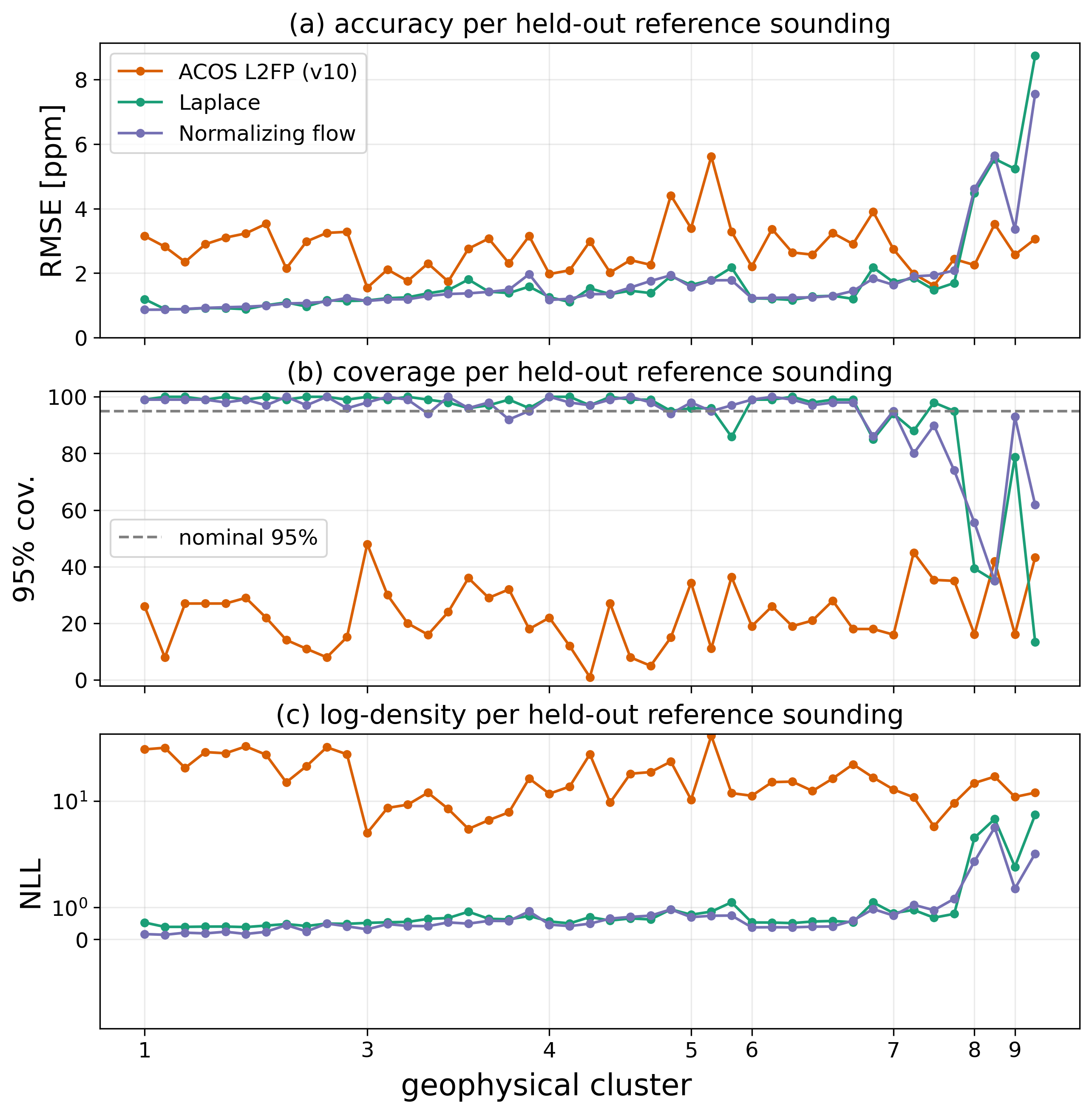}
    \caption{\textbf{Generalization to the 45 held-out reference soundings, XCO$_2$}. All three panels share one ordering: first geophysical cluster membership, then normalizing flow RMSE within each scene. \textbf{(a)} RMSE within each scene, ppm, logarithmic axis. The three methods track one another over most of the range and separate in the tail, which is the four scenes drawn from templates 8 and 9, the two that contribute fewest reference soundings to training (12 and 14 out of 361). On the other six templates the flow scores $1.40$ ppm against ACOS L2FP's $2.87$; on these four scenes it degrades to $5.5$ ppm while ACOS L2FP stays at $2.9$. The scenes are $8.8\%$ of the soundings and carry $60\%$ of the flow's total squared error. \textbf{(b)}. Empirical $95\%$ coverage within each scene; dashed line is nominal. The flow holds $93.4\%$ pooled and falls away on the tail on two of the four scenes. \textbf{(c)}. Mean NLL within each scene, standardized values.  }
    \label{fig:scenes_xco2}
\end{figure}

The degradation in accuracy is concentrated rather than general, and Figure \ref{fig:scenes_xco2} shows where. Two of the nine geophysical templates, clusters 8 and 9, contribute only $12$ and $14$ of the $361$ training reference soundings, and the four held-out scenes drawn from them ($395$ soundings, $8.8\%$ of the test set) carry $60\%$ of the flow's total squared XCO$_2$ error. Restricted to the other six templates the flow's XCO$_2$ RMSE is $1.40$ ppm against $2.87$ ppm for the operational retrieval and $1.76$ ppm for the prior XCO$_2$, so the accuracy advantage reported in Section \ref{sec:oco2_prediction} survives undiminished in the regimes the reference set samples adequately. The operational retrieval scores $2.87$ ppm on those six templates and $2.89$ ppm on the other two; the four scenes are therefore not harder for a retrieval that does not depend on training data. The prior XCO$_2$ errs by $5.15$~ppm on them, and the flow, at $5.5$~ppm, does no better than the prior there. The thin representation of their regime in training is a plausible explanation of the learned models' degradation, but these data do not separate it from other properties of these scenes.

\begin{figure}[htpb]
    \centering
    \includegraphics[width=\linewidth]{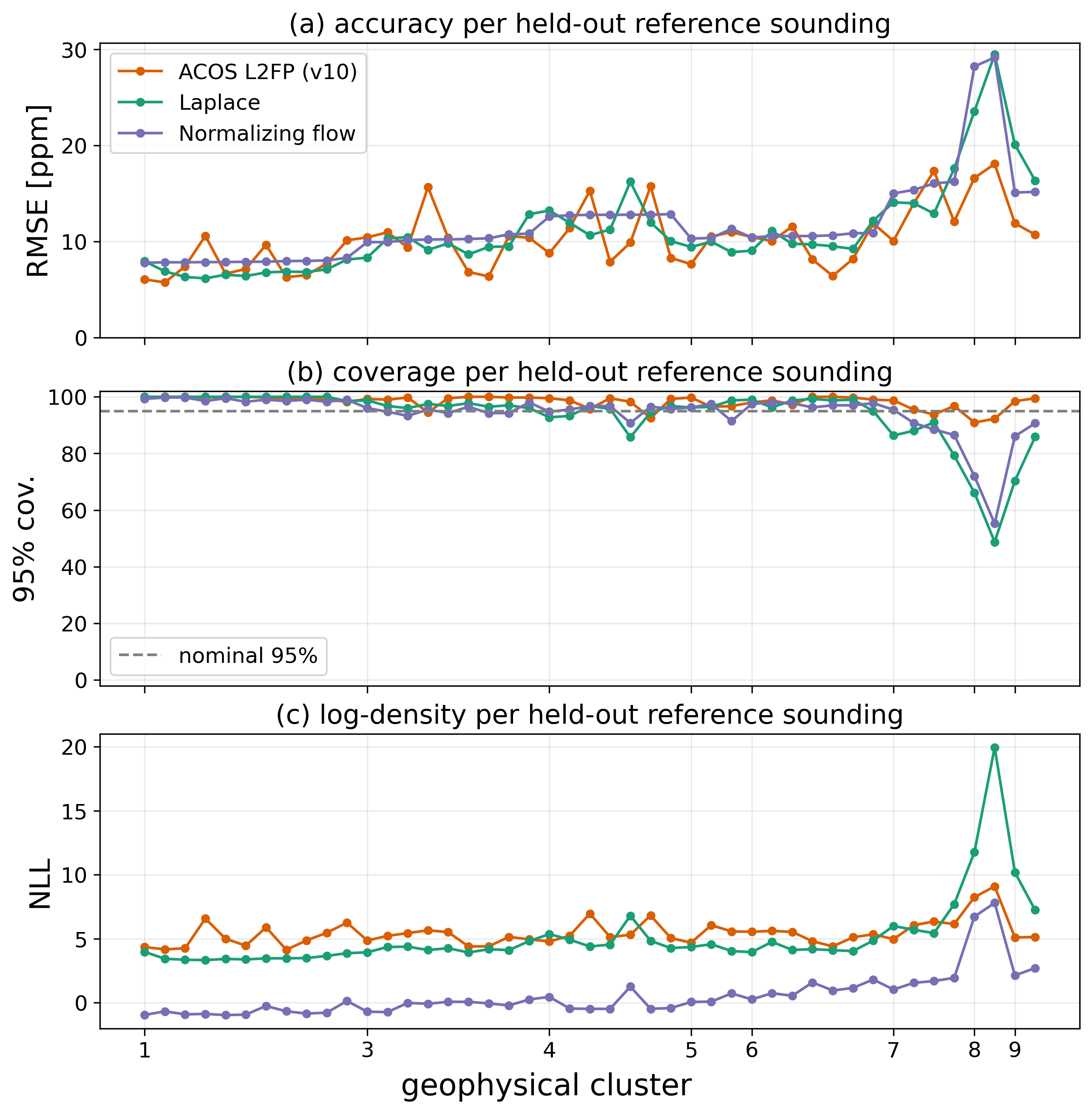}
    \caption{\textbf{Generalization to the $45$ held-out reference soundings, near-surface levels ($17$--$20$)}. Panels, ordering and tick labels as in Figure \ref{fig:scenes_xco2}. On the four scenes from template 8 and 9 the two learned models degrade to $23.0$ and $22.9$ ppm against ACOS L2FP's $14.7$. Pooled coverage is $94.3\%$ for the flow, $93.8\%$ for the Laplace approximation and $98.3\%$ for ACOS L2FP. In panel (c) the flow attains the lowest NLL in every one of the eight templates present, tail included: the advantage in predictive density survives where the advantage in accuracy does not. }
    \label{fig:scenes_last4}
\end{figure}

For the near-surface task see Figure \ref{fig:scenes_last4}. There the operational retrieval leads on the six well-represented templates as well ($10.19$ ppm against the flow's $10.95$), consistent with the observation in Section \ref{sec:last4} that the lowest pressure levels are where full-physics retrieval is most competitive. What survives on that task is the predictive density: the flow attains the lowest NLL in every one of the eight templates represented in the held-out partition, including the two sparse ones, where it scores $7.3$ against $8.7$ for the operational retrieval and $15.8$ for the Laplace approximation in cluster 8, and $2.4$ against $5.1$ and $8.8$ in cluster 9. A retrieval that is no more accurate than the operational one but attains a better predictive density is still very useful.

\begin{figure}
      \centering
      \includegraphics[width=\linewidth]{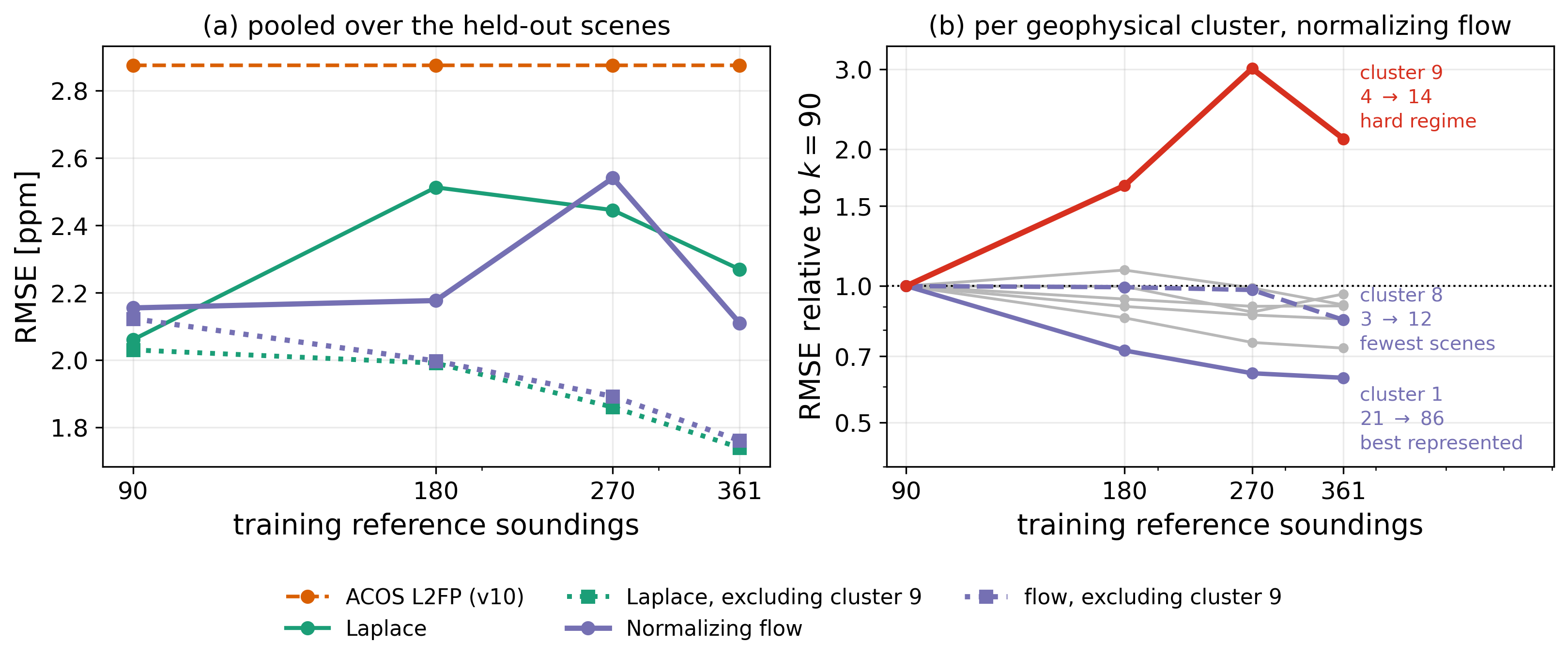}
      \caption{\textbf{Held-out error against the number of training reference soundings.} The scalar XCO$_2$ models retrained on nested, cluster-stratified subsets of the training partition and scored on the same $4{,}489$ held-out soundings. \textbf{(a)} Pooled over all $45$ held-out scenes neither learned method improves resolvably. Removing the two scenes of cluster $9$ (dotted) leaves both methods falling monotonically. \textbf{(b)} Per-cluster RMSE relative to $k=90$, logarithmic: seven of eight clusters improve while cluster $9$ degrades.}
      \label{fig:overfitting}
  \end{figure}

A plausible interpretation is that these results reflect the reference set more than the method: $451$ reference soundings spread unevenly over nine regimes is a small population from which to learn nine conditional posteriors. Figure \ref{fig:overfitting} is consistent with this reading: retrained on nested subsets of the training scenes, the flow improves in seven of the eight held-out regimes, and degrades on the eighth. It does not establish it, since nested subsets cannot add scenes from the sparse regimes.

To test whether the amortized retrieval transfers across time we obtained two further months of the same ensemble, March 2020 ($451$ reference soundings) and July 2020 ($448$), and evaluated on them without any fitting on the new months: the network weights and the target standardization are February's, and the Laplace approximation is calibrated on February's validation soundings. Because an unseen month makes February's own test partition unnecessary, the models evaluated here were retrained on all $406$ February reference soundings outside the validation set rather than the $361$ used above.

The two months are not equivalent tests. March lies almost entirely inside February's observed range: $2.9\%$ of its reference soundings fall outside February's range of scene-mean XCO$_2$ and $0.2\%$ outside its range of solar zenith angle. For July, XCO$_2$ scene-means  are $1.53$ ppm below February's and $30.4\%$ of its reference soundings have a solar zenith angle outside February's range. Since solar zenith angle is one of the network's inputs, July is a genuinely out-of-support test and March a same-season, in-support one.

Table~\ref{tab:months} reports both tasks, and it should be read against the ACOS L2FP row: the operational retrieval does not train, so its error on a new month measures the difficulty of that month rather than any model's ability to generalize. The flow remains the most accurate XCO$_2$ retrieval of the three in both months on all soundings, and it retains the best predictive density; its improvement over the prior XCO$_2$, however, is $8\%$ in RMSE in March and $2\%$ in July ($2.67$ against $2.73$~ppm). Near the surface the operational retrieval remains the most accurate, as it is on February data, and the two learned methods change places: the flow, behind the Laplace approximation on February by $0.53 \pm 0.32$ ppm, is ahead of it by $1.23 \pm 0.14$ ppm on March and $0.83 \pm 0.14$ ppm on July. Coverage does not transfer as well. Both learned methods undercover on both months, more so in July. For XCO$_2$ they remain far closer to nominal than ACOS L2FP, while for the near-surface target ACOS L2FP is closer to nominal than either, jointly and marginally.

On these simulations the prior XCO$_2$ is more accurate than the raw ACOS L2FP estimate ($2.22$ against $2.91$~ppm on the within-reference partition). The raw version-10 XCO$_2$ carries a global bias, which the operational bias correction removes and which the model discrepancy reproduces in the simulated retrievals: the mean error of ACOS L2FP is between $-1.9$ and $-1.6$~ppm in every setting of Tables~\ref{tab:splitcompare} and~\ref{tab:months}, against at most $0.4$~ppm in magnitude for the two learned methods. The simulated states were also only lightly filtered, which adds scatter. Removing each method's mean error lowers the ACOS L2FP RMSE to $2.24$~ppm within reference, and to $2.04$~ppm on soundings with a good quality flag, against $2.17$ and $1.97$~ppm for the prior and $1.46$ and $1.40$~ppm for the flow on the same soundings. With both corrections, ACOS L2FP is close to the flow on unseen reference soundings ($1.99$ against $1.95$~ppm) and in March ($2.18$ against $2.02$~ppm), and more accurate in July ($2.08$ against $2.57$~ppm); the flow's XCO$_2$ advantage is therefore largest where the training set represents the scene.

\subsection{Non-Gaussian posteriors}\label{sec:nongauss}

Operational retrieval theory relies on a locally Gaussian approximation to the posterior \citep{rodgers2000inverse}. It is worth being precise about where a departure from that approximation can come from in this experiment, because none of it is imposed. Given the state, every stochastic ingredient of the simulation is Gaussian: the discrepancy $\bfdelta|c \sim \normal(\tilde \bfmu_{c}, \tilde \bfSigma_{c})$ and the instrument noise $\bfepsilon \sim \normal(\bfzero, \bS_{\bfepsilon})$, drawn independently of each other and of the state, so that
\begin{equation}
    \by|\bx, c \sim \normal\left(\bF(\bx;\bb) + \tilde \bfmu_{c},\;
    \bS_{\bfepsilon} + \tilde \bfSigma_{c}\right).
\end{equation}
The posterior is nevertheless not Gaussian, for three reasons that are properties of the inverse problem rather than modeling choices. First, $\bF$ is nonlinear in $\bx$ \citep{odell2012ACOS, kulawik2019validation}. Second, the state ensemble $p_c(\bx)$ is a Gaussian mixture (Section \ref{subsec:dataset_comp}). Third, the cluster label is never an input, so the distribution the network is trained to approximate is the cluster-marginal
\begin{equation}
    p(\bx|\bu) = \sum_{c=1}^{9} \Pr(c|\bu)\, p(\bx|\bu, c), \qquad
    p(\bx|\bu, c) \propto \normal\left(\by; \bF(\bx;\bb) + \tilde \bfmu_{c},\;
    \bS_{\bfepsilon} + \tilde \bfSigma_{c}\right) p_c(\bx),
\end{equation}
a mixture of nine posteriors with different priors and different discrepancy moments; this is why the posterior geometry below varies systematically with cluster membership even though the label is withheld.

We evaluate whether amortized probabilistic retrieval with normalizing flows (NF) captures posterior features that are not representable in Gaussian approximations.

\begin{figure}[htpb]
    \centering
    \includegraphics[width=\linewidth]{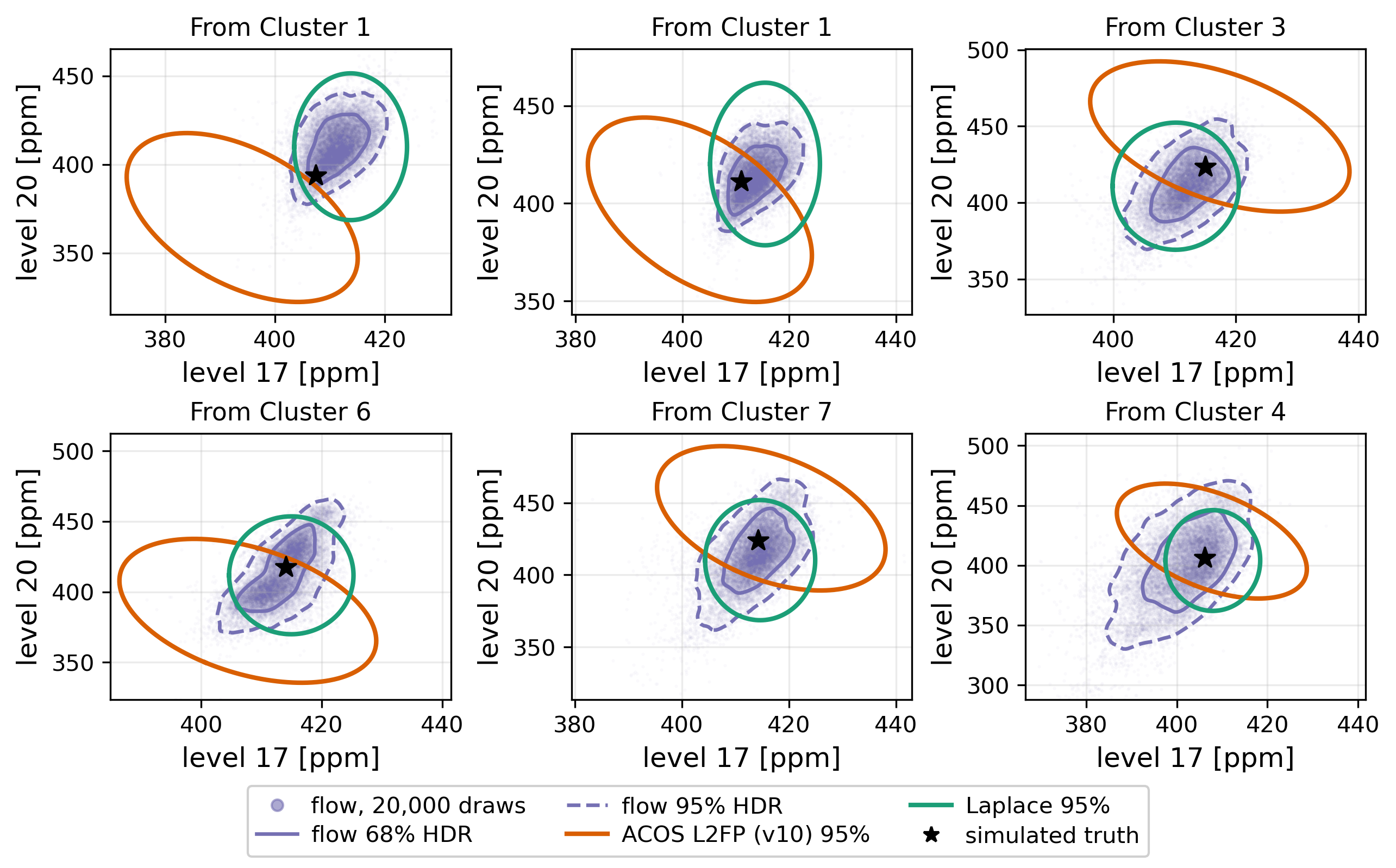}
    \caption{\textbf{Bivariate posteriors at $\rm{CO}_2$ levels $17$ and $20$ for six soundings drawn uniformly at random from the reference-sounding-disjoint test partition}. Points are $20{,}000$ draws from the normalizing flow's posterior and the violet curves its own $68\%$ (solid) and $95\%$ (dashed) highest-density regions (HDRs); the ellipses are the $95\%$ regions of the ACOS L2FP and Laplace posteriors projected onto the same two levels; the star is the simulated truth. The Gaussian regions are elliptical by construction while the flow's can be skewed and curved. The flow's $95\%$ regions are also markedly smaller.}
    \label{fig:bivar}
\end{figure}

Figure \ref{fig:bivar} shows bivariate posterior slices for two representative pressure levels (Level $20$, the surface, and level $17$, above the surface in the lower troposphere) for six randomly selected test soundings. Across many soundings, NF exhibits asymmetric and curved bivariate structures, as opposed to the Gaussian ellipses for the Laplace approximation and ACOS L2FP. Similar departures from normality in the $\rm{CO}_2$ profile posterior under OCO-2 have been documented by Markov Chain Monte Carlo (MCMC) comparisons to OE \citep{brynjarsdottir2018optimal,lamminpaa2019accelerated}. This is consistent with known physical non-linearities between the $\rm{CO}_2$ profile and the spectral measurements: The $\rm{O}_2$ band constraining path length and clouds/aerosols; strong $\rm{CO}_2$ having aerosol/water-vapor sensitivity, which can induce non-Gaussian ambiguity in the inverse problem \citep{odell2012ACOS, odell2018acos,lamminpaa2019accelerated}. Importantly, the network is never given the cluster label, yet the posterior geometry varies systematically with cluster membership, suggesting that posterior non-Gaussianity is driven by observable spectral geometry signatures; this aligns with the dataset design where clusters correspond to distinct geophysical regimes and discrepancy distributions. The Gaussian baselines also differ from the flow in scale: their projected $95\%$ regions are frequently much larger than the region the flow's samples occupy. Nevertheless, even at the nominal $95\%$ level, ACOS L2FP's Gaussian can miss the true simulated point (see, e.g., the top-right panel).

A curved contour is not by itself evidence that the curvature matters. To separate the departure from normality from the calibration of scale, we score each sounding under the Gaussian whose mean and covariance are those of the flow's own posterior.
The Gaussian's mean and covariance are Monte Carlo estimates from $S$ flow draws per sounding, so the two predictive distributions match on the first two moments up to Monte Carlo error, and any difference between them is shape beyond second order: it cannot be a better point estimate, and it cannot be a better-calibrated scale. Table \ref{tab:shape} reports the comparison at three values of $S$. A Gaussian fitted to $S$ draws pays an estimation penalty that shrinks as $S$ grows, so a small $S$ overstates the gap. For XCO$_2$ the gap does not depend on $S$; for the profile tasks it shrinks as $S$ grows, and at every $S$ it is more than twice its standard error. For profile predictions, the unconstrained flow is a much better alternative than its moment-matched Gaussian counterpart. It is also better for the scalar XCO$_2$ task, but the gap is smaller and not resolved by these data.

\begin{table}[htbp]
        \centering
        \begin{tabular}{lrrrrr}
        \toprule
         & & Moment-matched & \multicolumn{3}{c}{Shape gap (SE)} \\
        \cmidrule(lr){4-6}
        Task & Flow NLL & NLL ($S = 8{,}192$) & $S = 512$ & $S = 2{,}048$ & $S = 8{,}192$ \\
        \midrule
        XCO$_2$      & 0.736 & 0.882 & 0.148 (0.089) & 0.145 (0.087) & 0.146 (0.089) \\
        Near-surface & 0.526 & 1.059 & \textbf{0.602} (0.111) & \textbf{0.552} (0.096) & \textbf{0.533} (0.094) \\
        20-level     & $-8.319$ & $-3.554$ & \textbf{5.871} (0.749) & \textbf{5.116} (0.732) & \textbf{4.764} (0.735) \\
        \bottomrule
        \end{tabular}
        \caption{The conditional normalizing flow against the Gaussian matching its own posterior mean and covariance, sounding by sounding, on the reference-disjoint test partition ($n = 4{,}489$) defined in section \ref{subsec:group_split}. The Gaussian's mean and covariance are estimated from $S$ fresh flow draws per sounding, and the shape gap is reported at three values of $S$; the moment-matched NLL column is at $S = 8{,}192$. Because the two distributions match on the first two moments up to Monte Carlo error, the gap between them is attributable to posterior shape alone; positive values favor the flow. Bold marks gaps larger than twice their SE. Standard errors (SE) resample the held-out reference soundings.}
         \label{tab:shape}
        \end{table}

\subsection{Which bands carry the signal}\label{sec:occlusion}

A retrieval that reached its accuracy through the auxiliary covariates rather than through the spectra would be of little operational use. We test this directly: for one band at a time we replace the radiance channel by the mean spectrum of the training partition, leaving the wavelength channel and the two scale constants untouched, so the network still knows which band it is reading and on what scale and only the sounding-specific shape is removed. Table \ref{tab:attribution_xco2} reports the effect on the reference-disjoint XCO$_2$ task.

The weak $\rm{CO}_2$ band carries the point estimate for both models: occluding it costs $2.34$~ppm for the Laplace approximation and $3.40$~ppm for the flow, leaving both behind ACOS L2FP's $2.87$~ppm. The two models respond differently to the other bands. Occluding the strong $\rm{CO}_2$ band raises the Laplace approximation's RMSE by $0.46$~ppm but leaves the flow's unchanged within its standard error, and occluding all three bands costs the flow somewhat more than the Laplace approximation ($4.60$ against $3.57$~ppm, a margin comparable to their standard errors of $1.02$ and $0.59$). The two are trained on different objectives, squared error for the network underlying the Laplace approximation and conditional likelihood for the flow, and for XCO$_2$ the flow's context also bypasses the fusion module (Appendix~\ref{app:implementation}); either difference may lead them to weight the bands differently. The $\rm{O}_2$ A-band and the strong $\rm{CO}_2$ band leave the flow's point estimate unchanged within its standard error, yet each degrades its predictive density by a positive margin. Since RMSE depends only on the posterior mean, these bands inform the rest of the flow's predictive distribution, its spread, its shape, or both. This is consistent with their physical role: they constrain photon path length, aerosol and water vapor, which govern how well $\rm{CO}_2$ can be recovered more than how much of it there is.

The same substitution applied to the auxiliary inputs measures how much the fitted models rely on climatological information those inputs carry (Table~\ref{tab:attribution_xco2}, lower block). Replacing the a priori XCO$_2$ estimate by its training mean costs the flow $0.71$~ppm and the Laplace approximation $0.57$~ppm; replacing the observing geometry costs $2.43$ and $1.95$~ppm, consistent with its role in setting the optical path against which the spectra are interpreted \citep{odell2012ACOS}. Both are smaller than the cost of removing the spectra. The costs are not additive: replacing the prior and the geometry together costs less than replacing the geometry alone ($1.55$ against $2.43$~ppm for the flow, $1.48$ against $1.95$~ppm for the Laplace approximation), and the same holds with cluster means. Substituting a mean removes information but also creates combinations of inputs that never occur in training. Each row therefore measures the information removed together with the network's response to an unfamiliar input.

A network could still exploit climatology by recognizing the kind of scene from its inputs and returning the typical XCO$_2$ for it. Table~\ref{tab:scene_occlusion} tests this by repeating each occlusion with the mean of the sounding's own geophysical cluster, which preserves the information identifying the regime and removes only what is specific to the sounding. For the prior XCO$_2$ the cluster mean recovers about half of the cost of its removal ($0.71 \to 0.39$~ppm for the flow on unseen scenes), the regime-level component of its contribution. For the spectra it recovers about a third: the cluster's mean spectrum still costs the flow $2.97$~ppm on unseen scenes, so most of the spectral contribution requires the sounding's own spectrum. For the geometry it recovers nothing on unseen scenes ($2.43 \to 2.38$~ppm), so the network uses geometry per sounding rather than as a proxy for season or regime. On the within-reference partition the ordering is the same, with larger costs throughout.

\begin{table*}[htpb]
\centering
\small
\setlength{\tabcolsep}{4pt}
\renewcommand{\arraystretch}{1.15}
\begin{tabular}{l cc cc cc}
\toprule
 & \multicolumn{2}{c}{Laplace RMSE} & \multicolumn{2}{c}{NF RMSE} & \multicolumn{2}{c}{NF NLL} \\
\cmidrule(lr){2-3}\cmidrule(lr){4-5}\cmidrule(lr){6-7}
Occluded & value & $\Delta$ & value & $\Delta$ & value & $\Delta$ \\
\midrule
\multicolumn{7}{@{}l}{\textit{Spectral bands:}}\\
O$_2$            & 2.26 (0.37) & $+0.01$ (0.06) & 2.09 (0.30) & $-0.02$ (0.04) & 0.82 (0.11) & $\bf +0.08$ (0.02) \\
strong CO$_2$    & 2.71 (0.35) & $\bf +0.46$ (0.18) & 2.02 (0.24) & $-0.09$ (0.23) & 0.83 (0.12) & $\bf +0.10$ (0.04) \\
weak CO$_2$      & 4.59 (0.42) & $\bf +2.34$ (0.49) & 5.51 (0.82) & $\bf +3.40$ (0.78) & 4.31 (0.66) & $\bf +3.57$ (0.64) \\
all three        & 5.83 (0.48) & $\bf +3.57$ (0.59) & 6.71 (1.04) & $\bf +4.60$ (1.02) & 4.49 (0.72) & $\bf +3.75$ (0.70) \\
\midrule
\multicolumn{7}{@{}l}{\textit{Auxiliary inputs:}}\\
prior XCO$_2$    & 2.83 (0.37) & $\bf +0.57$ (0.20) & 2.82 (0.30) & $\bf +0.71$ (0.19) & 1.46 (0.15) & $\bf +0.72$ (0.14) \\
geometry         & 4.21 (0.27) & $\bf +1.95$ (0.37) & 4.54 (0.33) & $\bf +2.43$ (0.37) & 3.40 (0.43) & $\bf +2.66$ (0.42) \\
prior + geometry & 3.74 (0.26) & $\bf +1.48$ (0.33) & 3.67 (0.31) & $\bf +1.55$ (0.30) & 2.51 (0.32) & $\bf +1.77$ (0.32) \\
all continuous   & 4.39 (0.45) & $\bf +2.13$ (0.35) & 5.58 (0.89) & $\bf +3.47$ (0.71) & 3.68 (0.65) & $\bf +2.95$ (0.61) \\
\midrule
\multicolumn{7}{@{}l}{\textit{Reference values, not occluded:}}\\
Laplace / NF     & 2.26 (0.38) & -- & 2.11 (0.33) & -- & 0.74 (0.11) & -- \\
ACOS L2FP        & \multicolumn{4}{c}{2.87 (0.12)} & 20.20 (1.88) & -- \\
\bottomrule
\end{tabular}
\caption{\textbf{Input occlusion, XCO$_2$, reference-disjoint partition.} Parentheses are bootstrap standard errors resampling reference soundings. $\Delta$ columns measure difference in corresponding metric with the full, non-occluded method. Bolded values in those columns indicate the difference is at least two times bigger than the SE values. Spectral rows replace a band's radiance by a mean training spectrum. Auxiliary rows replace the named inputs by their training-set means: the prior XCO$_2$; the observing geometry (solar and viewing zenith angles, their secants, and air mass); both together; and all nine continuous auxiliary inputs, which adds the prior surface pressure and the two prior aerosol optical depths. Categorical inputs (footprint and aerosol types) are left unchanged.}
\label{tab:attribution_xco2}
\end{table*}

\begin{table*}[htpb]
\centering
\small
\setlength{\tabcolsep}{5pt}
\renewcommand{\arraystretch}{1.1}
\resizebox{\columnwidth}{!}{\begin{tabular}{l rr@{\hskip 14pt}rr}
\toprule
 & \multicolumn{2}{c}{Laplace $\Delta$RMSE} & \multicolumn{2}{c}{NF $\Delta$RMSE} \\
\cmidrule(lr){2-3}\cmidrule(lr){4-5}
Occluded & overall mean & cluster mean & overall mean & cluster mean \\
\midrule
\multicolumn{5}{l}{\emph{Reference-disjoint}} \\
\quad prior XCO$_2$   & $+0.57$ (0.20) & $+0.34$ (0.21) & $+0.71$ (0.19) & $+0.39$ (0.18) \\
\quad geometry        & $+1.95$ (0.37) & $+2.31$ (0.55) & $+2.43$ (0.37) & $+2.38$ (0.46) \\
\quad all three bands & $+3.57$ (0.59) & $+2.72$ (0.48) & $+4.60$ (1.02) & $+2.97$ (0.64) \\
\midrule
\multicolumn{5}{l}{\emph{Within-reference}} \\
\quad prior XCO$_2$   & $+1.00$ (0.07) & $+0.61$ (0.05) & $+1.12$ (0.07) & $+0.70$ (0.06) \\
\quad geometry        & $+3.55$ (0.13) & $+2.99$ (0.19) & $+3.51$ (0.12) & $+2.87$ (0.13) \\
\quad all three bands & $+7.18$ (0.32) & $+5.70$ (0.25) & $+6.06$ (0.26) & $+4.35$ (0.18) \\
\bottomrule
\end{tabular}
}
\caption{\textbf{Occlusion by overall and by cluster means, XCO$_2$.} Change in RMSE (ppm) when the named inputs are replaced by their training-set mean (overall) or by the training-set mean of the sounding's own geophysical cluster, relative to the intact model; inputs as in Table~\ref{tab:attribution_xco2}. The cluster mean keeps the information that identifies a sounding's regime and removes only the sounding-specific part, so the gap between the two columns bounds what a regime-level, climatological shortcut through that input could supply. Parentheses are paired bootstrap standard errors resampling reference soundings.}
\label{tab:scene_occlusion}
\end{table*}

\subsection{Computational efficiency at test time}

We compared the test-time computational cost of the retrieval methods considered in our study. Since our main interest is amortized inference post-training, we measure wall-clock prediction time on the test set. The results should be interpreted as the cost of generating a predictive distribution given some context as input.

We report median milliseconds per sounding at several batch sizes for the $20$-level CO$_2$ retrieval in Table \ref{tab:bench}. All timings are in single (FP32) precision, follow a discarded warm-up pass, and cover the predictive computation only: the standardization of inputs and the inverse transformation of outputs are excluded, while drawing the posterior is included. In practical terms, these results indicate a favorable accuracy--cost tradeoff. The Gaussian/Laplace baselines seem to be faster at bigger batch sizes, but the flow-based models do not incur a prohibitive inference penalty relative to the gains in calibration and likelihood-based performance reported above. This is particularly relevant in our setting, where the scientific value of the retrieval depends not only on point accuracy but also on a faithful characterization of posterior uncertainty. These computational costs are several orders of magnitude smaller than the operational ACOS L2FP retrieval, which comes at a cost of about 150 seconds per retrieval on average. That figure was measured while the simulation ensemble was generated, on different hardware and in a different software environment; we did not run the operational retrieval on the machine used for Table \ref{tab:bench}, so the two should be read as an order-of-magnitude contrast rather than a controlled benchmark. It is the most conservative estimate available to us: the literature reports comparable or larger costs, about  five minutes per sounding on a single processor \citep{taylor2023evaluating}, and $221$ s follows from the $55.26$ s per radiative-transfer evaluation and Jacobian of \citet{lamminpaa2025} at the median of four iterations reported in Section \ref{subsec:baselines}. It is important to note the differing natures of the methods: the trained models evaluate a fixed network, while the ACOS L2FP has to iterate a radiative-transfer forward model to convergence for every sounding, and it retrieves the whole state vector (which is of size $49$) and not just the elements of the state related to CO$_2$.

\begin{table}[htpb]
\centering
\begin{tabular}{lrrr}
\toprule
 & \multicolumn{3}{c}{ms per sounding} \\
\cmidrule(lr){2-4}
Method & batch 1 & batch 32 & batch 256 \\
\midrule
Point estimate & 2.05 & 0.064 & 0.008 \\
Laplace & 53.9 & 17.0 & 2.15 \\
Normalizing flow & 29.1 & 4.64 & 4.09 \\
\bottomrule
\end{tabular}
\caption{\textbf{Test-time cost per sounding for the $20$-level profile task}, on one NVIDIA A100 (SXM4, 80 GB)  (PyTorch 2.1.2, CUDA 12.1); median of five passes of eight batches reported. Laplace and Normalizing flow rows include time required to draw $512$ posterior samples.}
\label{tab:bench}
\end{table}

The timings in Table~\ref{tab:bench} measure post-training inference cost, not the cost of model selection. This distinction is important because training and hyperparameter search are offline costs that are amortized over future retrievals. Each model's search was $25$ Bayesian-optimization trials of $20$ epochs followed by a retrain of the selected configuration at $40$ epochs: a measured $5.5$ graphics processing unit (GPU) hours per model at the median, and $34.7$ GPU-hours for all six models reported here. On the other hand, inference cost can increase when many posterior samples are needed which require likelihood evaluations through both the context network and the spline-flow transformation. However, once training is complete, the resulting models can be evaluated at the millisecond-per-sounding scale. The efficiency comparison in this section therefore focuses on the online cost relevant for deployment, while acknowledging that the density estimators require a substantially larger one-time training investment than the deterministic baselines.

\section{Discussion and conclusions \label{sec:conclusions}}

\subsection{Interpretation: Bridging the gap between physics and reality}

The learned retrievals are more accurate than ACOS L2FP on the XCO$_2$ task. On February data they are also closer to nominal coverage than ACOS L2FP on all three tasks, except for the Laplace approximation's near-surface joint coverage. On the unseen months both undercover: for XCO$_2$ they remain far closer to nominal than ACOS L2FP, while for the near-surface target ACOS L2FP is closer, jointly and marginally. Operational full-physics retrieval solves an inverse problem in which the forward model $\bF$ is assumed to describe the radiative transfer exactly. When a systematic discrepancy $\bfdelta$ is present (imperfect spectroscopy, aerosol approximation, unmodeled behaviors) the solver adjusts the state $\bx$ to explain what $\bF$ cannot reproduce. The operational processing chain treats this with an empirical bias correction fitted against ground-based coincidences \citep{v10dug,odell2018acos}; the comparison in this work is against the raw state estimate.

In contrast, our networks minimize a loss in state space on pairs whose radiances contain $\bfdelta$, so the correction is learned internally rather than applied afterwards. It requires the discrepancy to be represented in the training distribution.

The two designs place similar corrections at different points in the pipeline, and that choice carries a cost and a statistical consequence. Applying it afterwards leaves the retrieval untouched when the correction is revised, which is what lets the operational chain update a data product without reprocessing: \citet{jacobs2024} delivered the v11.1 digital-elevation-model revision as a post-processing correction precisely because the L2FP retrieval, being computationally expensive, was not rerun. For an amortized retrieval, reprocessing the archive and refitting the model are not the different orders of expense that they are operationally.

The two designs also scale differently in the number of soundings $N$. Writing $C_F(N) = N t_F$ for the full-physics cost and $C_A(N) = T_0 + N t_N$ for the amortized one, with $T_0$ the one-time training cost, the measured values are $T_0 = 5.5$ GPU-hours per model (Section \ref{sec:experiments}), $t_N = 4.09$ ms per sounding for the $20$-level predictive with $512$ draws, and $t_F \approx 150$ s. The crossover is at $N^{*} \approx 132$ soundings, under two minutes of quality-passing data from an instrument that collects on the order of one million soundings per day. In the most expensive scenario, where a change to the forward model or the spectroscopy forces the ensemble itself to be regenerated, the $44{,}982$ simulated soundings at the $33.45$ s per forward evaluation reported by \citet{lamminpaa2025} come to roughly $420$ core-hours, and the crossover moves to about $10{,}000$ soundings. Retraining is therefore not the operational obstacle it might appear to be, provided the need for it can be detected.

The normalizing flow attained the best predictive density (lowest NLL) in every scenario considered in this work, but advantage in point prediction was not achieved in all scenarios. If we use a grouped (by reference sounding) train-test split, or evaluate the learned February model on other held-out months, ACOS L2FP produces more accurate point estimates than our methods in certain scenarios (especially, in retrieving near-surface CO$_2$ quantities). The evidence of Section~\ref{subsec:group_split} is consistent with limited diversity of the training scenes as a cause, but does not establish it: further experiments in future work are in order.

\subsection{The value of simulation-based training}
The success of this approach is inextricably linked to the high fidelity of the training data. In real-world remote sensing, ground truth is sparse, limited to a few dozen TCCON sites globally, and even at these sites the variable measured by TCCON is not the same as the one targeted by OCO-2 due to different spatial resolutions. This makes it difficult to train deep, data-hungry models without overfitting to localized climatologies (e.g., learning that ``latitude Z implies $CO_2$ X'').

By utilizing the Jet Propulsion Laboratory's OCO-2 UQ simulation dataset, we trained our models in an environment where the truth is known everywhere. This lets the network be trained without geographic or temporal inputs, so that it cannot read location or date directly. It still has access to climatological information: the a priori XCO$_2$ estimate is climatological by construction, and the spectra and observing geometry carry scene properties that can correlate with season and location. The occlusion experiments of Section~\ref{sec:occlusion} assess this reliance. Removing the a priori XCO$_2$ estimate costs the flow $0.71$~ppm on unseen scenes against $4.60$ for removing the spectra, and the point estimate is carried by the weak CO$_2$ band rather than by the bands that encode surface, aerosol and water vapor. The ability of the Normalizing Flows to capture non-Gaussian posteriors demonstrates that the network can learn a non-linear topology related to the solution manifold, shown by the superior NLL performance of this method, plausibly identifying the ``null space'' where different atmospheric states produce identical spectra.

What the simulation dataset does not provide is a large population of distinct scenes. The $451$ reference soundings are spread unevenly over geophysical regimes, as discussed in Section \ref{subsec:group_split}. Two regimes contribute only $12$ and $14$ reference soundings to training, and the four held-out reference soundings drawn from them ($8.8\%$ of the test partition) carry $60\%$ of the flow's total squared XCO$_2$ error. ACOS L2FP, which does not train, is no worse on those scenes than elsewhere.

\subsection{Limitations and future directions}
While our results are promising, we acknowledge the ``Synthetic Gap.'' Our model is currently trained and evaluated entirely on simulated data. While the simulations are designed to be rigorous, real-world instrument noise and environmental heterogeneity may exhibit distributions that differ from our training set.

The immediate next step is to bridge this gap through domain adaptation. Future work will involve fine-tuning these pre-trained networks on real coincident measurements from TCCON. Furthermore, while our method provides a ``global'' correction for model error, it assumes the error structure in the real world mirrors the discrepancy $\boldsymbol{\delta}$ modeled in the simulations. Continuous validation against real-world data remains essential to ensure that the learned bias corrections remain valid across different seasons and geographies.

    The moments for the discrepancy $\bfdelta$ are constant within clusters, which reflects our belief that the discrepancy between ``full-physics'' models and reality incurs in similar kinds of errors for similar geophysical states. This constant distribution per cluster is a reasonable assumption, but an obvious refinement would be to model $\bfdelta(\bs)$ as a smooth field making use of geographic information, which only enters the current model through the observing geometry of the reference sounding.

We trained all our models on a February 2020 simulated dataset, and validated on two further months. Thus, our results show promising evidence that data-driven retrievals via deep learning can be accurate and calibrated in this limited scenario. An important future direction is to obtain reference soundings at differing seasons for training diversity, but also to include temporal information directly into the model, perhaps as another auxiliary covariate that would help the model predict. This will also help us assess the generalizability of our model across seasons, years and under temporal distribution shift. Seasonal sampling of the training set is a recurring concern for learned XCO$_2$ retrievals. \citet{xie2024} sample OCO-2 soundings across the four seasons and report that their network reproduces the seasonal cycle observed at TCCON sites, and \citet{reuter2025} train on alternate weeks of a simulated year so that seasonal variations are finely sampled, yet still find errors that vary through the year. Our February-only models are the limiting case of a training set that does not sample the seasonal cycle. In our setting the season also enters through the inputs: $30.4\%$ of July's reference soundings have a solar zenith angle outside February's range. Solar zenith and azimuth angles can be strong proxies for season and location \citep[Chapter~2]{keely2025dissertation}. Because our networks are trained on a single month they cannot have learned geometry as a seasonal proxy, and the occlusion experiments of Section~\ref{sec:occlusion} indicate that they use it per sounding rather than as a regime-level proxy (Table~\ref{tab:scene_occlusion}). A model trained across seasons could learn such a proxy, and the same occlusion tests would be one way to check for it.

The February dataset contains land-only simulations at only $451$ locations world-wide. To assess the full geographic generalizability potential of the model a denser, more diverse (geographically) sample of reference soundings must be considered.

Amortization moves the computation up front, and that cost is paid once per algorithm version: a revision to the system's components might warrant a refitting of the network. This depends on what component changes: a geolocation correction changes the values of the covariates the network reads rather than the mapping between them and the state, so inference can simply be re-run, provided the corrected values remain inside the training ensemble's support. A known deterministic change to the L1b convention can be applied to the training radiance and the models refitted without new radiative transfer. A change to the forward model or the spectroscopy requires the simulation ensemble to be regenerated and discrepancy $\bfdelta$ to be re-estimated. What an operational deployment would also need is a monitor to assess the need for re-training: two-sample tests on the encoder's own summary statistics against the training distribution \citep{rabanser2019failing, schmitt2024detecting}; the per-band spectral residual $\chi^2$ evaluated at the network's posterior mean, which is the statistic the discrepancy was estimated from \citep{braverman2021post}; and a small full-physics audit fraction, which at $1\%$ of the quality-passing data stream costs $1\%$ of what processing that stream costs and yields a continuous comparison of the two retrievals on identical inputs. Refitting would then be triggered by those statistics rather than by the calendar \citep{gama2014,kirkpatrick2017overcoming,parisi2019continual}. An amortized retrieval in operations is thus a maintained system under monitoring rather than a fixed artifact \citep{sculley2015hidden}.

As with other deep neural network methods, we lack mechanistic interpretability and we cannot claim to know which layers/features of the network encode the CO$_2$ signal. Some work has been done in this avenue for other atmospheric retrieval problems \citep{yip2021} but such an analysis is out of the scope of this work.

We found that profile accuracy is not better on our learned models with respect to ACOS L2FP on unseen scenes, and that joint calibration degrades: on the $20$-level task under the reference-disjoint partition no method is jointly calibrated, the observed $95\%$ joint rates being $44.3\%$, $81.7\%$ and $58.7\%$ for ACOS L2FP, the Laplace approximation and the flow respectively (Table \ref{tab:splitcompare}). We report this as a limitation of all three methods on that task rather than of the learned ones alone, and the evidence of Section~\ref{subsec:group_split} suggests, but does not establish, that this could be improved via better training-set diversity, which points to an avenue for future work.

We note that the ACOS L2FP is one configuration of the method, not the full operational product with bias-correction. Future work should be done to compare against the latest ACOS L2FP configuration used operationally, with bias-correction mechanisms.

\subsection{Conclusion}
In this work, we presented a novel framework for the amortized probabilistic retrieval of atmospheric carbon dioxide, using spectral and auxiliary encoders which are then fused to produce a coherent representation of the data, which are then used by two posterior approximations: a Laplace approximation over the last layer and a conditional normalizing flow.

Our contributions are threefold. First, we demonstrated that amortized inference, once trained, reduces retrieval time from minutes to milliseconds on the hardware described in Section \ref{sec:experiments}, and that calibrated posteriors are attainable at that cost: for XCO$_2$, coverage is close to nominal on February data
and, although it degrades on unseen months, remains much closer to nominal than the operational retrieval's. Second, we showed that by training on simulations with explicit model discrepancy, our approach yields more accurate point estimates than the raw output of traditional OE on the XCO$_2$ setup, while maintaining competitive accuracies on the profile setups, and is nearly unbiased where the raw retrieval is not; once the raw retrieval's mean error is removed and its quality filter applied, the XCO$_2$ advantage remains within reference but is small or reversed out of sample (Section(Section~\ref{subsec:group_split}). Third, the normalizing flow's advantage in predictive density is attained even when accuracy advantages are not: it attains the best likelihood on all tasks and partitions; per geophysical regime, it attains the lowest NLL on all eight regimes present in the held-out partition and for the two out-of-sample months, on both profiles and the scalar task, including both regimes the reference set barely samples. A retrieval with competitive (but worse) RMSE and better predictive density is a valuable tool, although lower NLL does not by itself establish calibration. Everything above also applies to OCO-3: it measures the same bands and its radiances are processed with ACOS L2FP \citep{basilio2019orbiting,taylor2020}, so the encoder, discrepancy and posteriors carry over. What does not carry is the training set: viewing geometries across missions might differ, and a matching OCO-3 reference ensemble would have to be constructed for training.

   Taken together, these results point to the number and spread of distinct scenes available for training as a plausible main constraint on amortized retrieval in this setting. $451$ reference soundings over nine regimes, two of them sampled a dozen times, is a small population from which to learn nine conditional posteriors.
   There are plausible solutions to this problem: enlarging the population so that it contains a more diverse set of reference soundings, and applying standard remedies against sparsely populated regions of the input space, such as density-based weighting of the loss \citep{shimodaira2000,steininger2021,yang2021}. These changes, plus validation against TCCON data, are the paths to a retrieval that is faster, more accurate and better calibrated than current baselines, and have thus great potential as operational tools.

\appendix
\section*{Acknowledgments}

The authors were partially supported by NASA's Advanced Information Systems Technology Program (AIST-21). MK's research was partially supported by the Office of the Vice Chancellor for Research at the University of Wisconsin--Madison with funding from the Wisconsin Alumni Research Foundation. Part of this work was performed at the Jet Propulsion Laboratory, California Institute of Technology, under contract with NASA.
We would like to thank Amy Braverman and Otto Lamminp\"{a}\"{a} for helpful comments and discussions.

\section{Performance Metrics}\label{sec:metrics}

All metrics are evaluated on the held-out test set $\{(\mathbf{u}_i, \mathbf{x}_i)\}_{i=1}^{n_\text{test}}$, where $\mathbf{u}_i$ denotes the full spectral and auxiliary input and $\mathbf{x}_i \in \mathbb{R}^D$ denotes the true geophysical state. For profile prediction tasks ($D = 20$ or $D = 4$), we index individual pressure levels by $d = 1, \ldots, D$. Let $\hat{\mu}_i = \hat{\mu}(\mathbf{u}_i)$ denote the predictive mean and $\hat{\Sigma}_{\text{pred},i} = \hat{\Sigma}_\text{pred}(\mathbf{u}_i)$ denote the predictive covariance for observation $i$.

\paragraph*{Root Mean Squared Error (RMSE)}

measures point prediction accuracy. For scalar XCO$_2$ prediction ($D = 1$):
\begin{equation}
    \text{RMSE} = \sqrt{\frac{1}{n_\text{test}} \sum_{i=1}^{n_\text{test}} (x_i - \hat{\mu}_i)^2}
\end{equation}
For profile prediction ($D > 1$), we report RMSEs pooled over soundings and levels:

\begin{equation}
    \text{RMSE} =  \sqrt{\frac{1}{n_\text{test} D} \sum_{i=1}^{n_{\text{test}}}\sum_{d=1}^{D } (x_{id} - \hat{\mu}_{id})^2}
\end{equation}
where $x_{id}$ and $\hat{\mu}_{id}$ denote the $d$-th component of $\mathbf{x}_i$ and $\hat{\mu}_i$, respectively. RMSE is reported in parts per million (ppm).

\paragraph*{Mean Absolute Error (MAE)}
summarizes the same residuals as the RMSE under a loss that is linear rather than quadratic in the error. Writing $\tilde{x}_{id}$ for the predictive median of component $d$ for sounding $i$ at which a method is judged,
\begin{equation*}
    \text{MAE} = \frac{1}{n_\text{test} D}
      \sum_{i=1}^{n_\text{test}} \sum_{d=1}^{D} \left| x_{id} - \tilde{x}_{id} \right|,
\end{equation*}
pooled over soundings and pressure levels and reported in ppm. We report it alongside
the RMSE because the residuals here are heavy tailed: a small fraction of soundings
carries a disproportionate share of the squared error, so the RMSE is to a large extent
a statement about that tail while the MAE describes the typical sounding. When the two
order the methods differently, the difference between them is localized to the bulk or
to the tail rather than being absorbed into a single number.

The two metrics are minimized by different point estimates: the RMSE by the predictive mean, the MAE by the predictive median. Each method is evaluated at its own MAE-optimal estimate. For ACOS L2FP and for the Laplace approximation the predictive is Gaussian and the two coincide, so the MAE column uses the same $\hat{\mu}_i$ as the RMSE column. For the normalizing flow, whose predictive need not be symmetric, we use the componentwise empirical median of the same $512$ posterior draws used for the marginal coverages.

\paragraph*{Negative Log-Likelihood (NLL)}
evaluates the full predictive distribution, penalizing both inaccurate point estimates and miscalibrated uncertainty. For each test observation, let $p(\mathbf{x}_i \mid \mathbf{u}_i)$ denote the predictive density evaluated at the true state---either a Gaussian $\mathcal{N}(\mathbf{x}_i;\, \hat{\mu}_i, \hat{\Sigma}_{\text{pred},i})$ for the Laplace approximation, or the normalizing flow density. The average NLL over the test set is:
\begin{equation}
    \text{NLL} = -\frac{1}{n_\text{test}} \sum_{i=1}^{n_\text{test}} \log p(\mathbf{x}_i \mid \mathbf{u}_i)
\end{equation}
Lower NLL indicates a better-fitting predictive distribution. For profile tasks, we report the joint NLL (over all $D$ levels simultaneously). All NLL values reported in this paper are joint predictive densities, evaluated in the standardized target coordinates in which the models are trained, $x_{\mathrm{std}, d} = (x_d - \mu_d)/\sigma_d$, with $\mu_d$ and $\sigma_d$ estimated from training soundings alone. The change of variables to parts per million is affine, so
\begin{equation*}
    \mathrm{NLL}_{\mathrm{ppm}} = \mathrm{NLL}_\mathrm{std} + \sum_{d=1}^D \log \left( \sigma_d \cdot 10^6 \right)
\end{equation*}
which equals $1.14, 9.60$, and $32.50$ for the scalar, near-surface and $20$-level targets respectively. The correction is identical for every method and every sounding, so it changes no ranking within a table.

Most tables compare methods on one target, one partition and one set of soundings. NLL values are \emph{not} comparable across the scalar, near-surface and $20$-level tasks in either coordinate system: a joint density over $D$ levels is not comparable to a density over one level. The comparison above removes the standardization but not the dimension.

\paragraph*{Empirical Coverage}

measures whether a predictive distribution's credible regions contain the truth at the advertised rate. For a nominal level $\alpha \in (0,1)$ and a region $\mathcal{R}_{\alpha,i} \subseteq \mathbb{R}^{D}$ constructed from the predictive for sounding $i$, we report the empirical rate over the test set,
\begin{equation*}
    \widehat{\text{cov}}(\alpha)
      = \frac{1}{n_\text{test}} \sum_{i=1}^{n_\text{test}}
        \mathbf{1}\!\left(\mathbf{x}_i \in \mathcal{R}_{\alpha,i}\right),
\end{equation*}
which estimates $P(\mathbf{x} \in \mathcal{R}_{\alpha}(\mathbf{u}))$ averaged over the test population. A calibrated method gives $\widehat{\text{cov}}(\alpha) \approx \alpha$; below is under-coverage, above is over-coverage, and neither direction is preferable.

For $D > 1$ the choice of $\mathcal{R}_{\alpha,i}$ is not unique, and we report two quantities that answer different questions.

\emph{Marginal coverage} takes $\mathcal{R}_{\alpha,i}$ to be the central $\alpha$ interval of a single pressure level $d$ and averages the resulting rate over the $D$ levels. For the two Gaussian predictives the interval is $\hat{\mu}_{id} \pm \Phi^{-1}\!\left(\tfrac{1+\alpha}{2}\right) \hat{\Sigma}_{\text{pred},i,dd}^{1/2}$; for the normalizing flow it is the pair of empirical $\tfrac{1-\alpha}{2}$ and $\tfrac{1+\alpha}{2}$ quantiles of $512$ posterior draws, so no Gaussian assumption is imposed on it. Marginal coverage tests each level's predictive separately and is by construction insensitive to the dependence between levels.

\emph{Joint coverage} takes $\mathcal{R}_{\alpha,i}$ to be a region of the full $D$-dimensional predictive carrying probability $\alpha$ under that predictive. For a Gaussian predictive this is the HDR, an ellipsoid that in terms of the whitened residual $z_i = L_i^{-1}(\mathbf{x}_i - \hat{\mu}_i)$, with $\hat{\Sigma}_{\text{pred},i} = L_i L_i^{\top}$ the Cholesky factorization, is the ball
\begin{equation}
    \|z_i\|^{2} \;\le\; \chi^{2}_{D}(\alpha).
    \label{eq:jointcov}
\end{equation}
For the normalizing flow we use the same region in its own latent space: the flow maps the target to a standard normal base density through $z_i = f_{x}(\mathbf{x}_i \mid \mathbf{u}_i)$, so \eqref{eq:jointcov} defines the pre-image of the standard normal region of probability $\alpha$, which by the change of variables carries probability exactly $\alpha$ under the flow's predictive. This region is an exact $\alpha$-region of the flow's predictive but is not in general its HDR, since the Jacobian of $f_x$ is not constant; we use it because it is exact, cheap and defined identically for every sounding, and we refer to it as the joint region rather than as an HDR throughout. At $D = 1$ the two definitions coincide and a single number is reported.

\section{Implementation details}\label{app:implementation}

This appendix gives the implementation details for the neural regressors and the conditional normalizing flows used. All models take as input the three OCO-2 spectral bands together with the auxiliary covariates described in Subsection \ref{subsec:dataset_comp}. The neural regressors share a common encoder structure, but differ in the output dimension and in whether the final model is a point predictor or a conditional density estimator.

The spectral encoder processes the three spectral bands separately. For each band, the input is a two-channel one-dimensional signal, together with a two-dimensional vector of band-level constants. The spectral signal is passed through four one-dimensional convolutional blocks with channel sizes
\[
    2 \to 4 \to 8 \to 16 \to 32,
\]
where each block uses a kernel size of \(3\), stride \(2\), padding \(1\), batch normalization, and a ReLU activation. The resulting feature map is passed through a max-pooling layer with kernel size \(2\) and stride \(2\), flattened, concatenated with the two band-level constants, and then mapped to an \(E\)-dimensional vector by a two-layer MLP with one hidden ReLU activation. This gives one \(E\)-dimensional token for each of the three spectral bands.

The auxiliary covariates are encoded separately. The nine continuous covariates are standardized and mapped jointly to a $72$-dimensional vector by a single linear layer, and each of the three categorical covariates is mapped to an 8-dimensional learned embedding: the footprint index, with eight observed levels and one reserved for a level not seen in training, and the two aerosol types, with four observed levels each and one reserved. The resulting $96$-dimensional auxiliary representation is linearly projected to dimension \(E\). The full encoder therefore returns four \(E\)-dimensional tokens: one for each spectral band and one for the auxiliary covariates. The encoder output has shape $4 \times E$.

The attention-fusion module first applies layer normalization to these four tokens and then passes them through \(L\) Transformer encoder layers with \(H\) attention heads and embedding dimension \(E\). The resulting sequence of four transformed tokens is flattened to obtain a \(4E\)-dimensional representation.

In the deterministic setup, the fused representation is passed through a three-layer MLP, i.e.,

\begin{equation*}
    \mathbb{R}^{4E}
    \longrightarrow
    \mathbb{R}^{E}
    \longrightarrow
    \mathbb{R}^{E}
    \longrightarrow
    \mathbb{R}^{D}
\end{equation*}

with ReLU activations after two linear layers. The output dimension $D$ depends on the target, $D=1$ for scalar XCO$_2$, $D= 20$ for the full profile, and $D=4$ for the last-four near-surface profile.

The deterministic regressors are trained by minimizing MSE on the selected target coordinates. We use AdamW \citep{loshchilov2017decoupled} with a cosine-annealing learning-rate schedule.

For scalar XCO$_2$, the conditional normalizing flow uses the scalar backbone encoder as a context network. In this case, we do not use the attention-fusion output as context; instead, we flatten the encoder output to obtain a context vector of dimension \(4E\). The scalar density estimator is a one-dimensional conditional masked rational-quadratic autoregressive spline flow with linear tails. Each spline transformation is followed by a random permutation, which is a no-op in one dimension but keeps the implementation consistent with the multivariate flows. The base distribution is a standard normal distribution on \(\mathbb{R}\).

For the profile-valued density estimators, we use the profile backbone as the context network. In these models, the context is the representation obtained after the encoder and the attention-fusion module. Thus, for embedding dimension \(E\), the context dimension is \(4E\). The profile flows are conditional rational-quadratic coupling flows. The coupling layers alternate complementary binary masks and include random permutations between layers to mix coordinates. The conditioner in each coupling layer is a residual network with two residual blocks and ReLU activations. The base distribution is a standard normal distribution on \(\mathbb{R}^D\), where \(D=20\) for the full profile and \(D=4\) for the near-surface model.

All flow models are trained by minimizing the negative conditional log-likelihood. The backbone encoder is trained jointly with the flow weights. For the scalar flow, the encoder learning rate is set to one quarter of the flow learning rate for stability.

\section{Hyperparameter search and model selection}

The deterministic regressors are selected by validation RMSE. The conditional normalizing flows are selected by validation NLL. For all three tasks, we use Optuna \citep{akiba2019optuna} searching over learning rate, weight decay, flow depth, hidden width, number of spline bins, and tail bound. The weights for the NF backbone are randomly initialized before training (i.e., we do not take into account the learned weights from MSE training, and train using NLL from scratch). Table~\ref{tab:model_hyperparams} summarizes the final configurations used in the experiments.

\begin{table*}[htpb]
\centering
\resizebox{\textwidth}{!}{
\begin{tabular}{llcccclcccccl}
\toprule
Model & Target & \(D\) & \(E\) & \(H\) & \(L\) & Flow & Flow layers & Hidden & Bins
& Tail bound & Batch & Optimizer \\
\midrule

Deterministic & XCO$_2$ & 1 & 320 & 4 & 4 & -- & -- & -- & -- & -- & 512 & lr $=1.06\times10^{-4}$, wd $=2.32\times10^{-5}$ \\

Deterministic & Full profile & 20 & 256 & 4 & 1 & -- & -- & -- & -- & -- & 512 & lr $=7.99\times10^{-4}$, wd $=1.82\times10^{-3}$ \\

Deterministic & Near-surface & 4 & 256 & 4 & 1 & -- & -- & -- & -- & -- & 256 & lr $=3.60\times10^{-4}$, wd $=2.51\times10^{-6}$ \\

NF & XCO$_2$ & 1 & 512 & -- & -- & AR RQ spline & 10 & 744 & 7 & 6.52 & 128 & lr $=4.03\times10^{-4}$, wd $=1.03\times10^{-6}$ \\

NF & Full profile & 20 & 128 & 8 & 1 & Coupling RQ spline & 12 & 727 & 10 & 9.09 & 128 & lr $=5.04\times10^{-4}$, wd $=1.75\times10^{-5}$ \\

NF & Near-surface & 4 & 384 & 4 & 3 & Coupling RQ spline & 11 & 674 & 8 & 5.38 & 128 & lr $=7.55\times10^{-4}$, wd $=3.75\times10^{-6}$ \\

\bottomrule
\end{tabular}
}
\caption{Final model configurations. \(D\) is the output dimension, \(E\) the embedding dimension, \(H\) the number of attention heads and \(L\) the number of Transformer layers in the attention-fusion module; the flow columns give the number of spline transforms, the width of each conditioner, the number of bins and the tail bound. The scalar flow conditions on the encoder output without the fusion module (Appendix~\ref{app:implementation}), so \(H\) and \(L\) do not apply to it. AR RQ spline is an autoregressive rational-quadratic spline flow and Coupling RQ spline a rational-quadratic coupling flow; lr and wd are the learning rate and weight decay. Every configuration is the winner of a $25$-trial Bayesian-optimization search over hyperparameters, retrained from scratch for 40 epochs at the selected configuration. Deterministic regressors were selected on validation RMSE and the flows on validation NLL. All models use AdamW with cosine-annealed
learning rate.}
\label{tab:model_hyperparams}
\end{table*}

\section*{Open Research Statement}

Code for reproducing results described here is attached as a zip and can be found at the GitHub repository \url{https://github.com/acallesalda/oco2-amortized-uq}, with persistent code record at \url{https://zenodo.org/records/24025972}, while data, split tables, predictive artifacts and trained model weights can be found at \url{https://zenodo.org/records/24025972}.

\section*{Conflicts of Interest}

The authors declare there are no conflicts of interest for this manuscript.

\bibliographystyle{apalike}
\bibliography{mendeley,additionalrefs}

@Article{eldering2017orbiting,
AUTHOR = {Eldering, A. and O'Dell, C. W. and Wennberg, P. O. and Crisp, D. and Gunson, M. R. and Viatte, C. and Avis, C. and Braverman, A. and Castano, R. and Chang, A. and Chapsky, L. and Cheng, C. and Connor, B. and Dang, L. and Doran, G. and Fisher, B. and Frankenberg, C. and Fu, D. and Granat, R. and Hobbs, J. and Lee, R. A. M. and Mandrake, L. and McDuffie, J. and Miller, C. E. and Myers, V. and Natraj, V. and O'Brien, D. and Osterman, G. B. and Oyafuso, F. and Payne, V. H. and Pollock, H. R. and Polonsky, I. and Roehl, C. M. and Rosenberg, R. and Schwandner, F. and Smyth, M. and Tang, V. and Taylor, T. E. and To, C. and Wunch, D. and Yoshimizu, J.},
TITLE = {The {Orbiting Carbon Observatory}-2: first 18~months of science data products},
JOURNAL = {Atmospheric Measurement Techniques},
VOLUME = {10},
YEAR = {2017},
NUMBER = {2},
PAGES = {549--563},
URL = {https://amt.copernicus.org/articles/10/549/2017/},
DOI = {10.5194/amt-10-549-2017}
}

@article{arrhenius1896xxxi,
  title={On the influence of carbonic acid in the air upon the temperature of the ground},
  author={Arrhenius, Svante},
  journal={The London, Edinburgh, and Dublin Philosophical Magazine and Journal of Science},
  volume={41},
  number={251},
  pages={237--276},
  year={1896},
  publisher={Taylor \& Francis}
}

@article{braverman2021post,
  title={Post hoc uncertainty quantification for remote sensing observing systems},
  author={Braverman, Amy and Hobbs, Jonathan and Teixeira, Joaquim and Gunson, Michael},
  journal={SIAM/ASA Journal on Uncertainty Quantification},
  volume={9},
  number={3},
  pages={1064--1093},
  year={2021},
  publisher={SIAM}
}

@inproceedings{basilio2019orbiting,
  title={{Orbiting Carbon Observatory-3 (OCO-3)}, remote sensing from the {International Space station (ISS)}},
  author={Basilio, Ralph R and Bennett, Matthew W and Eldering, Annmarie and Lawson, Peter R and Rosenberg, Robert A},
  booktitle={Sensors, Systems, and Next-Generation Satellites XXIII},
  volume={11151},
  pages={42--55},
  year={2019},
  organization={SPIE}
}

@article{patil2022objective,
  title={Objective Frequentist Uncertainty Quantification for Atmospheric Retrievals},
  author={Patil, Pratik and Kuusela, Mikael and Hobbs, Jonathan},
  journal={SIAM/ASA Journal on Uncertainty Quantification},
  volume={10},
  number={3},
  pages={827--859},
  year={2022},
  publisher={SIAM}
}

@misc{ocoatbd_v10,
    author = {Crisp, D. and O'Dell, C. and Eldering, A. and Fisher, B. and Oyafuso, F. and Payne, V. and Drouin, B. and Toon, G. and Laughner, J. and Somkuti, P. and McGarragh, G. and Merrelli, A. and Nelson, R. and Gunson, M. and Frankenberg, C. and Osterman, G. and Boesch, H. and Brown, L. and Castano, R. and Christi, M. and Connor, B. and McDuffie, J. and Miller, C. and Natraj, V. and O'Brien, D. and Polonski, I. and Smyth, M. and Thompson, D. and Granat, R.}, 
    title = {{Orbiting Carbon Observatory-2 \& 3 ({OCO-2} \& {OCO-3}) Level 2 Full Physics Retrieval
Algorithm Theoretical Basis, Version 3.0}},
    howpublished = {URL: \url{https://docserver.gesdisc.eosdis.nasa.gov/public/project/OCO/OCO_L2_ATBD.pdf}},
    year = {2021}
}

@article{david2021xco,
  title={ {XCO}$_2$ estimates from the {OCO-2} measurements using a neural network approach},
  author={David, Leslie and Br{\'e}on, Fran{\c{c}}ois-Marie and Chevallier, Fr{\'e}d{\'e}ric},
  journal={Atmospheric Measurement Techniques},
  volume={14},
  number={1},
  pages={117--132},
  year={2021},
  publisher={Copernicus Publications G{\"o}ttingen, Germany}
}

@article{breon2022potential,
  title={On the potential of a neural-network-based approach for estimating {XCO}$_2$ from {OCO-2} measurements},
  author={Br{\'e}on, Fran{\c{c}}ois-Marie and David, Leslie and Chatelanaz, Pierre and Chevallier, Fr{\'e}d{\'e}ric},
  journal={Atmospheric Measurement Techniques},
  volume={15},
  number={18},
  pages={5219--5234},
  year={2022},
  publisher={Copernicus Publications G{\"o}ttingen, Germany}
}

@article{keely2023nonlinear,
  title={A nonlinear data-driven approach to bias correction of {XCO}$_2$ for {NASA's OCO-2 ACOS} version 10},
  author={Keely, William R and Mauceri, Steffen and Crowell, Sean and O'Dell, Christopher W},
  journal={Atmospheric Measurement Techniques},
  volume={16},
  number={23},
  pages={5725--5748},
  year={2023},
  publisher={Copernicus Publications G{\"o}ttingen, Germany}
}

@article{keely2025conditional,
  title={Conditional Diffusion-Based Retrieval of Atmospheric {CO}$_2$ from {Earth Observing Spectroscopy}},
  author={Keely, William R and Lamminp{\"a}{\"a}, Otto and Mauceri, Steffen and Crowell, Sean MR and O'Dell, Christopher W and McGarragh, Gregory R},
  journal={arXiv preprint arXiv:2504.17074},
  year={2025}
}

@article{chen2025deterministic,
  title={From Deterministic to Probabilistic: A Lightweight Framework for Probabilistic Machine Learning in Trace Gas Remote Sensing},
  author={Chen, Wei and Ren, Tao and Zhao, Changying},
  journal={Journal of Remote Sensing},
  volume={5},
  pages={0881},
  year={2025},
  publisher={AAAS}
}

@article{chen2025transformer,
  title={Transformer-Based Fast Mole Fraction of {CO}$_2$ Retrievals from Satellite-Measured Spectra},
  author={Chen, Wei and Ren, Tao and Zhao, Changying and Wen, Yuan and Gu, Yilei and Zhou, Minqiang and Wang, Pucai},
  journal={Journal of Remote Sensing},
  volume={5},
  pages={0470},
  year={2025},
  publisher={AAAS}
}

@Article{odell2012ACOS,
AUTHOR = {O'Dell, C. W. and Connor, B. and B\"osch, H. and O'Brien, D. and Frankenberg, C. and Castano, R. and Christi, M. and Eldering, D. and Fisher, B. and Gunson, M. and McDuffie, J. and Miller, C. E. and Natraj, V. and Oyafuso, F. and Polonsky, I. and Smyth, M. and Taylor, T. and Toon, G. C. and Wennberg, P. O. and Wunch, D.},
TITLE = {The {ACOS} {CO}$_{2}$ retrieval algorithm – {Part 1}: {Description} and validation against synthetic observations},
JOURNAL = {Atmospheric Measurement Techniques},
VOLUME = {5},
YEAR = {2012},
NUMBER = {1},
PAGES = {99--121},
URL = {https://amt.copernicus.org/articles/5/99/2012/},
DOI = {10.5194/amt-5-99-2012}
}

@Article{crisp2012ACOS,
AUTHOR = {Crisp, D. and Fisher, B. M. and O'Dell, C. and Frankenberg, C. and Basilio, R. and B\"osch, H. and Brown, L. R. and Castano, R. and Connor, B. and Deutscher, N. M. and Eldering, A. and Griffith, D. and Gunson, M. and Kuze, A. and Mandrake, L. and McDuffie, J. and Messerschmidt, J. and Miller, C. E. and Morino, I. and Natraj, V. and Notholt, J. and O'Brien, D. M. and Oyafuso, F. and Polonsky, I. and Robinson, J. and Salawitch, R. and Sherlock, V. and Smyth, M. and Suto, H. and Taylor, T. E. and Thompson, D. R. and Wennberg, P. O. and Wunch, D. and Yung, Y. L.},
TITLE = {The {ACOS} {CO}$_2$ retrieval algorithm; {Part II}: {Global} {XCO}$_2$ data characterization},
JOURNAL = {Atmospheric Measurement Techniques},
VOLUME = {5},
YEAR = {2012},
NUMBER = {4},
PAGES = {687--707},
URL = {https://amt.copernicus.org/articles/5/687/2012/},
DOI = {10.5194/amt-5-687-2012}
}

@Article{connor2016UQ,
AUTHOR = {Connor, B. and B\"osch, H. and McDuffie, J. and Taylor, T. and Fu, D. and Frankenberg, C. and O'Dell, C. and Payne, V. H. and Gunson, M. and Pollock, R. and Hobbs, J. and Oyafuso, F. and Jiang, Y.},
TITLE = {Quantification of uncertainties in {OCO-2} measurements of {XCO}$_{2}$:
simulations and linear error analysis},
JOURNAL = {Atmospheric Measurement Techniques},
VOLUME = {9},
YEAR = {2016},
NUMBER = {10},
PAGES = {5227--5238},
URL = {https://amt.copernicus.org/articles/9/5227/2016/},
DOI = {10.5194/amt-9-5227-2016}
}

@Article{kulawik2019validation,
AUTHOR = {Kulawik, S. S. and O'Dell, C. and Nelson, R. R. and Taylor, T. E.},
TITLE = {Validation of {OCO-2} error analysis using simulated retrievals},
JOURNAL = {Atmospheric Measurement Techniques},
VOLUME = {12},
YEAR = {2019},
NUMBER = {10},
PAGES = {5317--5334},
URL = {https://amt.copernicus.org/articles/12/5317/2019/},
DOI = {10.5194/amt-12-5317-2019}
}

@article{hobbs2017simulation,
  title={Simulation-based uncertainty quantification for estimating atmospheric {CO}$_2$ from satellite data},
  author={Hobbs, Jonathan and Braverman, Amy and Cressie, Noel and Granat, Robert and Gunson, Michael},
  journal={SIAM/ASA Journal on Uncertainty Quantification},
  volume={5},
  number={1},
  pages={956--985},
  year={2017},
  publisher={SIAM}
}

@article{hobbs2020spectroscopic,
  title={Spectroscopic uncertainty impacts on {OCO-2/3} retrievals of {XCO}$_2$},
  author={Hobbs, J. and Drouin, Brian J and Oyafuso, Fabiano and Payne, Vivienne H and Gunson, Michael R and McDuffie, James and Mlawer, Eli J},
  journal={Journal of Quantitative Spectroscopy and Radiative Transfer},
  volume={257},
  pages={107360},
  year={2020},
  publisher={Elsevier}
}

@Article{taylor2023evaluating,
AUTHOR = {Taylor, T. E. and O'Dell, C. W. and Baker, D. and Bruegge, C. and Chang, A. and Chapsky, L. and Chatterjee, A. and Cheng, C. and Chevallier, F. and Crisp, D. and Dang, L. and Drouin, B. and Eldering, A. and Feng, L. and Fisher, B. and Fu, D. and Gunson, M. and Haemmerle, V. and Keller, G. R. and Kiel, M. and Kuai, L. and Kurosu, T. and Lambert, A. and Laughner, J. and Lee, R. and Liu, J. and Mandrake, L. and Marchetti, Y. and McGarragh, G. and Merrelli, A. and Nelson, R. R. and Osterman, G. and Oyafuso, F. and Palmer, P. I. and Payne, V. H. and Rosenberg, R. and Somkuti, P. and Spiers, G. and To, C. and Weir, B. and Wennberg, P. O. and Yu, S. and Zong, J.},
TITLE = {Evaluating the consistency between {OCO}-2 and {OCO}-3 {XCO}$_2$ estimates derived from the {NASA ACOS} version 10 retrieval algorithm},
JOURNAL = {Atmospheric Measurement Techniques},
VOLUME = {16},
YEAR = {2023},
NUMBER = {12},
PAGES = {3173--3209},
URL = {https://amt.copernicus.org/articles/16/3173/2023/},
DOI = {10.5194/amt-16-3173-2023}
}

@book{rodgers2000inverse,
  title={Inverse methods for atmospheric sounding: theory and practice},
  author={Rodgers, Clive D},
  volume={2},
  year={2000},
  publisher={World scientific}
}

@inproceedings{laplace2021,
  title={Laplace Redux--Effortless {B}ayesian Deep Learning},
  author={Erik Daxberger and Agustinus Kristiadi and Alexander Immer
          and Runa Eschenhagen and Matthias Bauer and Philipp Hennig},
  booktitle={{N}eur{IPS}},
  year={2021}
}

@Article{odell2018acos,
AUTHOR = {O'Dell, C. W. and Eldering, A. and Wennberg, P. O. and Crisp, D. and Gunson, M. R. and Fisher, B. and Frankenberg, C. and Kiel, M. and Lindqvist, H. and Mandrake, L. and Merrelli, A. and Natraj, V. and Nelson, R. R. and Osterman, G. B. and Payne, V. H. and Taylor, T. E. and Wunch, D. and Drouin, B. J. and Oyafuso, F. and Chang, A. and McDuffie, J. and Smyth, M. and Baker, D. F. and Basu, S. and Chevallier, F. and Crowell, S. M. R. and Feng, L. and Palmer, P. I. and Dubey, M. and Garc\'{\i}a, O. E. and Griffith, D. W. T. and Hase, F. and Iraci, L. T. and Kivi, R. and Morino, I. and Notholt, J. and Ohyama, H. and Petri, C. and Roehl, C. M. and Sha, M. K. and Strong, K. and Sussmann, R. and Te, Y. and Uchino, O. and Velazco, V. A.},
TITLE = {Improved retrievals of carbon dioxide from {Orbiting Carbon Observatory-2} with the version 8 {ACOS} algorithm},
JOURNAL = {Atmospheric Measurement Techniques},
VOLUME = {11},
YEAR = {2018},
NUMBER = {12},
PAGES = {6539--6576},
URL = {https://amt.copernicus.org/articles/11/6539/2018/},
DOI = {10.5194/amt-11-6539-2018}
}

@article{brynjarsdottir2018optimal,
  title={Optimal Estimation Versus {MCMC} for {CO$_2$} Retrievals},
  author={Brynjarsdottir, Jenny and Hobbs, Jonathan and Braverman, Amy and Mandrake, Lukas},
  journal={Journal of Agricultural, Biological and Environmental Statistics},
  volume={23},
  number={2},
  pages={297--316},
  year={2018},
  publisher={Springer}
}

@article{lamminpaa2019accelerated,
  title={Accelerated {MCMC} for satellite-based measurements of atmospheric {CO}$_2$},
  author={Lamminp{\"a}{\"a}, Otto and Hobbs, Jonathan and Brynjarsd{\'o}ttir, Jenn{\`y} and Laine, Marko and Braverman, Amy and Lindqvist, Hannakaisa and Tamminen, Johanna},
  journal={Remote Sensing},
  volume={11},
  number={17},
  pages={2061},
  year={2019},
  publisher={MDPI}
}

@inproceedings{nflows2020,
 author = {Durkan, Conor and Bekasov, Artur and Murray, Iain and Papamakarios, George},
 booktitle = {Advances in Neural Information Processing Systems},
 editor = {H. Wallach and H. Larochelle and A. Beygelzimer and F. d\textquotesingle Alch\'{e}-Buc and E. Fox and R. Garnett},
 pages = {},
 publisher = {Curran Associates, Inc.},
 title = {Neural Spline Flows},
 url = {https://proceedings.neurips.cc/paper_files/paper/2019/file/7ac71d433f282034e088473244df8c02-Paper.pdf},
 volume = {32},
 year = {2019}
}

@article{payne2020,
title = {Absorption coefficient ({ABSCO}) tables for the Orbiting Carbon Observatories: Version 5.1},
journal = {Journal of Quantitative Spectroscopy and Radiative Transfer},
volume = {255},
pages = {107217},
year = {2020},
issn = {0022-4073},
doi = {https://doi.org/10.1016/j.jqsrt.2020.107217},
url = {https://www.sciencedirect.com/science/article/pii/S0022407320302016},
author = {Vivienne H. Payne and Brian J. Drouin and Fabiano Oyafuso and Le Kuai and Brendan M. Fisher and Keeyoon Sung and Deacon Nemchick and Timothy J. Crawford and Mike Smyth and David Crisp and Erin Adkins and Joseph T. Hodges and David A. Long and Eli J. Mlawer and Aronne Merrelli and Elizabeth Lunny and Christopher W. O’Dell}
}

@article{kendall2017uncertainties,
  title={What uncertainties do we need in {Bayesian} deep learning for computer vision?},
  author={Kendall, Alex and Gal, Yarin},
  journal={Advances in neural information processing systems},
  volume={30},
  year={2017}
}

@article{hullermeier2021aleatoric,
  title={Aleatoric and epistemic uncertainty in machine learning: An introduction to concepts and methods},
  author={H{\"u}llermeier, Eyke and Waegeman, Willem},
  journal={Machine learning},
  volume={110},
  number={3},
  pages={457--506},
  year={2021},
  publisher={Springer}
}

@article{loshchilov2017decoupled,
  title={Decoupled weight decay regularization},
  author={Loshchilov, Ilya and Hutter, Frank},
  journal={arXiv preprint arXiv:1711.05101},
  year={2017}
}

@inproceedings{akiba2019optuna,
  title={{O}ptuna: A Next-Generation Hyperparameter Optimization Framework},
  author={Akiba, Takuya and Sano, Shotaro and Yanase, Toshihiko and Ohta, Takeru and Koyama, Masanori},
  booktitle={The 25th ACM SIGKDD International Conference on Knowledge Discovery \& Data Mining},
  pages={2623--2631},
  year={2019}
}

@article{mendonca2021assessing,
  title={Assessing the feasibility of using a neural network to filter {Orbiting Carbon Observatory 2 (OCO-2)} retrievals at northern high latitudes},
  author={Mendonca, Joseph and Nassar, Ray and O'Dell, Christopher W and Kivi, Rigel and Morino, Isamu and Notholt, Justus and Petri, Christof and Strong, Kimberly and Wunch, Debra},
  journal={Atmospheric Measurement Techniques},
  volume={14},
  number={12},
  pages={7511--7524},
  year={2021},
  publisher={Copernicus Publications G{\"o}ttingen, Germany}
}

@article{li2023influence,
  title={The Influence of Validation Colocation on {XCO}$_2$ Satellite--Terrestrial Joint Observations},
  author={Li, Ruoxi and Zhou, Xiang and Cheng, Tianhai and Tao, Zui and Gu, Xingfa and Wang, Ning and Zhang, Hongming and Lv, Tingting},
  journal={Remote Sensing},
  volume={15},
  number={22},
  pages={5270},
  year={2023},
  publisher={MDPI}
}

@article{miller_precision_2007,
	title = {Precision requirements for space-based {XCO}$_2$ data},
	volume = {112},
	doi = {10.1029/2006JD007659},
	number = {D10},
	journal = {Journal of Geophysical Research: Atmospheres},
	author = {Miller, C. E. and Crisp, D. and DeCola, P. L. and Olsen, S. C. and Randerson, J. T. and Michalak, A. M. and Alkhaled, A. and Rayner, P. and Jacob, D. J. and Suntharalingam, P. and Jones, D. B. A. and Denning, A. S. and Nicholls, M. E. and Doney, S. C. and Pawson, S. and Boesch, H. and Connor, B. J. and Fung, I. Y. and O'Brien, D. and Salawitch, R. J. and Sander, S. P. and Sen, B. and Tans, P. and Toon, G. C. and Wennberg, P. O. and Wofsy, S. C. and Yung, Y. L. and Law, R. M.},
	year = {2007},
	pages = {2006JD007659}
}

@article{nguyen2019,
  title={Sensitivity of Optimal Estimation Satellite Retrievals to Misspecification of the Prior Mean and Covariance, with Application to {OCO-2} Retrievals},
  author={Nguyen, Hai and Cressie, Noel and Hobbs, Jonathan},
  journal={Remote Sensing},
  volume={11},
  number={23},
  pages={2770},
  year={2019},
  doi = {10.3390/rs11232770}
}

@misc{ipcc_ar6_syr_2023,
  author = {{IPCC}},
  title = {Climate {Change} 2023: {Synthesis} {Report}. {Contribution} of {Working} {Groups} {I}, {II} and {III} to the {Sixth} {Assessment} {Report} of the {Intergovernmental} {Panel} on {Climate} {Change} {[Core Writing Team, H. Lee and J. Romero (eds.)]}},
  year = {2023},
  institution = {Intergovernmental Panel on Climate Change (IPCC)},
  address = {Geneva, Switzerland},
  publisher = {IPCC, Geneva, Switzerland},
  doi = {10.59327/IPCC/AR6-9789291691647},
  url = {https://www.ipcc.ch/report/ar6/syr/},
  editor = {Core Writing Team, H. Lee and J. Romero (eds.)},
  pages = {184}
}

@misc{v10dug,
  title = {{Orbiting Carbon Observatory-2 \& 3 Data Product User's Guide, Operational Level 2 Data Versions 10 and Lite File Version 10 and VEarly}},
  author={Osterman, G. and O'Dell, C. and Eldering, A. and Fisher, B. and Crisp, D. and Cheng, C. and Frankenberg, C. and Lambert, A. and Gunson, M. and Mandrake, L. and Wunch, D.},
  year={2020},
  publisher={Pasadena, CA},
  howpublished = {URL: \url{https://docserver.gesdisc.eosdis.nasa.gov/public/project/OCO/OCO2\_OCO3\_B10\_DUG.pdf}}
}

@Article{laughner2024total,
AUTHOR = {Laughner, J. L. and Toon, G. C. and Mendonca, J. and Petri, C. and Roche, S. and Wunch, D. and Blavier, J.-F. and Griffith, D. W. T. and Heikkinen, P. and Keeling, R. F. and Kiel, M. and Kivi, R. and Roehl, C. M. and Stephens, B. B. and Baier, B. C. and Chen, H. and Choi, Y. and Deutscher, N. M. and DiGangi, J. P. and Gross, J. and Herkommer, B. and Jeseck, P. and Laemmel, T. and Lan, X. and McGee, E. and McKain, K. and Miller, J. and Morino, I. and Notholt, J. and Ohyama, H. and Pollard, D. F. and Rettinger, M. and Riris, H. and Rousogenous, C. and Sha, M. K. and Shiomi, K. and Strong, K. and Sussmann, R. and T\'e, Y. and Velazco, V. A. and Wofsy, S. C. and Zhou, M. and Wennberg, P. O.},
TITLE = {The Total Carbon Column Observing Network's GGG2020 data version},
JOURNAL = {Earth System Science Data},
VOLUME = {16},
YEAR = {2024},
NUMBER = {5},
PAGES = {2197--2260},
URL = {https://essd.copernicus.org/articles/16/2197/2024/},
DOI = {10.5194/essd-16-2197-2024}
}

@misc{yadav2025error,
      title={An Error Model for Evaluating the Accuracy of Satellite-Based {XCO}$_2$ Products}, 
      author={Vineet Yadav and Jonathan Hobbs and Hai M. Nguyen and Susan S. Kulawik and Junjie Liu and David F. Baker and Isamu Morino and Hirofumi Ohyama and Voltaire A. Velazco and Mihalis Vrekoussis and Manvendra K. Dubey},
      year={2025},
      eprint={2509.16419},
      archivePrefix={arXiv},
      primaryClass={stat.AP},
      url={https://arxiv.org/abs/2509.16419}, 
}

@Article{xie2024,
AUTHOR = {Xie, F. and Ren, T. and Zhao, C. and Wen, Y. and Gu, Y. and Zhou, M. and Wang, P. and Shiomi, K. and Morino, I.},
TITLE = {Fast retrieval of {XCO}$_2$ over east Asia based on Orbiting Carbon Observatory-2 ({OCO-2}) spectral measurements},
JOURNAL = {Atmospheric Measurement Techniques},
VOLUME = {17},
YEAR = {2024},
NUMBER = {13},
PAGES = {3949--3967},
URL = {https://amt.copernicus.org/articles/17/3949/2024/},
DOI = {10.5194/amt-17-3949-2024}
}

@article{houweling2015,
author = {Houweling, S. and Baker, D. and Basu, S. and Boesch, H. and Butz, A. and Chevallier, F. and Deng, F. and Dlugokencky, E. J. and Feng, L. and Ganshin, A. and Hasekamp, O. and Jones, D. and Maksyutov, S. and Marshall, J. and Oda, T. and O'Dell, C. W. and Oshchepkov, S. and Palmer, P. I. and Peylin, P. and Poussi, Z. and Reum, F. and Takagi, H. and Yoshida, Y. and Zhuravlev, R.},
title = {An intercomparison of inverse models for estimating sources and sinks of {CO}2 using {GOSAT} measurements},
journal = {Journal of Geophysical Research: Atmospheres},
volume = {120},
number = {10},
pages = {5253-5266},
doi = {https://doi.org/10.1002/2014JD022962},
url = {https://agupubs.onlinelibrary.wiley.com/doi/abs/10.1002/2014JD022962},
eprint = {https://agupubs.onlinelibrary.wiley.com/doi/pdf/10.1002/2014JD022962},
year = {2015}
}

@Article{chevallier2019,
AUTHOR = {Chevallier, F. and Remaud, M. and O'Dell, C. W. and Baker, D. and Peylin, P. and Cozic, A.},
TITLE = {Objective evaluation of surface- and satellite-driven carbon dioxide atmospheric inversions},
JOURNAL = {Atmospheric Chemistry and Physics},
VOLUME = {19},
YEAR = {2019},
NUMBER = {22},
PAGES = {14233--14251},
URL = {https://acp.copernicus.org/articles/19/14233/2019/},
DOI = {10.5194/acp-19-14233-2019}
}

@Article{crowell2019,
AUTHOR = {Crowell, S. and Baker, D. and Schuh, A. and Basu, S. and Jacobson, A. R. and Chevallier, F. and Liu, J. and Deng, F. and Feng, L. and McKain, K. and Chatterjee, A. and Miller, J. B. and Stephens, B. B. and Eldering, A. and Crisp, D. and Schimel, D. and Nassar, R. and O'Dell, C. W. and Oda, T. and Sweeney, C. and Palmer, P. I. and Jones, D. B. A.},
TITLE = {The 2015--2016 carbon cycle as seen from {OCO-2} and the global in situ network},
JOURNAL = {Atmospheric Chemistry and Physics},
VOLUME = {19},
YEAR = {2019},
NUMBER = {15},
PAGES = {9797--9831},
URL = {https://acp.copernicus.org/articles/19/9797/2019/},
DOI = {10.5194/acp-19-9797-2019}
}

@Article{lamminpaa2025,
AUTHOR = {Lamminp\"a\"a, O. and Susiluoto, J. and Hobbs, J. and McDuffie, J. and Braverman, A. and Owhadi, H.},
TITLE = {Forward model emulator for atmospheric radiative transfer using {Gaussian} processes and cross validation},
JOURNAL = {Atmospheric Measurement Techniques},
VOLUME = {18},
YEAR = {2025},
NUMBER = {3},
PAGES = {673--694},
URL = {https://amt.copernicus.org/articles/18/673/2025/},
DOI = {10.5194/amt-18-673-2025}
}

@Article{jacobs2024,
AUTHOR = {Jacobs, N. and O'Dell, C. W. and Taylor, T. E. and Logan, T. L. and Byrne, B. and Kiel, M. and Kivi, R. and Heikkinen, P. and Merrelli, A. and Payne, V. H. and Chatterjee, A.},
TITLE = {The importance of digital elevation model accuracy in {XCO}$_2$ retrievals: improving the {Orbiting Carbon Observatory} 2 Atmospheric Carbon Observations from Space version 11 retrieval product},
JOURNAL = {Atmospheric Measurement Techniques},
VOLUME = {17},
YEAR = {2024},
NUMBER = {5},
PAGES = {1375--1401},
URL = {https://amt.copernicus.org/articles/17/1375/2024/},
DOI = {10.5194/amt-17-1375-2024}
}

@article{taylor2020,
title = {{OCO}-3 early mission operations and initial (vEarly) {XCO}$_2$ and {SIF} retrievals},
journal = {Remote Sensing of Environment},
volume = {251},
pages = {112032},
year = {2020},
issn = {0034-4257},
doi = {https://doi.org/10.1016/j.rse.2020.112032},
url = {https://www.sciencedirect.com/science/article/pii/S0034425720304028},
author = {Thomas E. Taylor and Annmarie Eldering and Aronne Merrelli and Matthäus Kiel and Peter Somkuti and Cecilia Cheng and Robert Rosenberg and Brendan Fisher and David Crisp and Ralph Basilio and Matthew Bennett and Daniel Cervantes and Albert Chang and Lan Dang and Christian Frankenberg and Vance R. Haemmerle and Graziela R. Keller and Thomas Kurosu and Joshua L. Laughner and Richard Lee and Yuliya Marchetti and Robert R. Nelson and Christopher W. O'Dell and Gregory Osterman and Ryan Pavlick and Coleen Roehl and Robert Schneider and Gary Spiers and Cathy To and Christopher Wells and Paul O. Wennberg and Amruta Yelamanchili and Shanshan Yu}
}

@Article{nguyen2020,
AUTHOR = {Nguyen, Hai and Hobbs, Jonathan},
TITLE = {Intercomparison of Remote Sensing Retrievals: An Examination of Prior-Induced Biases in Averaging Kernel Corrections},
JOURNAL = {Remote Sensing},
VOLUME = {12},
YEAR = {2020},
NUMBER = {19},
ARTICLE-NUMBER = {3239},
URL = {https://www.mdpi.com/2072-4292/12/19/3239},
ISSN = {2072-4292},
DOI = {10.3390/rs12193239}
}

@article{wunch2011,
    author = {Wunch, Debra and Toon, Geoffrey C. and Blavier, Jean-François L. and Washenfelder, Rebecca A. and Notholt, Justus and Connor, Brian J. and Griffith, David W. T. and Sherlock, Vanessa and Wennberg, Paul O.},
    title = {The Total Carbon Column Observing Network},
    journal = {Philosophical Transactions of the Royal Society A: Mathematical, Physical and Engineering Sciences},
    volume = {369},
    number = {1943},
    pages = {2087-2112},
    year = {2011},
    month = {05},
    issn = {1364-503X},
    doi = {10.1098/rsta.2010.0240},
    url = {https://doi.org/10.1098/rsta.2010.0240},
    eprint = {https://royalsocietypublishing.org/rsta/article-pdf/369/1943/2087/1348887/rsta.2010.0240.pdf},
}

@article{
chatterjee2017,
author = {A. Chatterjee  and M. M. Gierach  and A. J. Sutton  and R. A. Feely  and D. Crisp  and A. Eldering  and M. R. Gunson  and C. W. O’Dell  and B. B. Stephens  and D. S. Schimel },
title = {Influence of {El Niño} on atmospheric {CO}$_2$ over the tropical {Pacific Ocean}: Findings from {NASA}'s {OCO}-2 mission},
journal = {Science},
volume = {358},
number = {6360},
pages = {eaam5776},
year = {2017},
doi = {10.1126/science.aam5776},
URL = {https://www.science.org/doi/abs/10.1126/science.aam5776},
eprint = {https://www.science.org/doi/pdf/10.1126/science.aam5776}}

@article{
schwander2017,
author = {Florian M. Schwandner  and Michael R. Gunson  and Charles E. Miller  and Simon A. Carn  and Annmarie Eldering  and Thomas Krings  and Kristal R. Verhulst  and David S. Schimel  and Hai M. Nguyen  and David Crisp  and Christopher W. O’Dell  and Gregory B. Osterman  and Laura T. Iraci  and James R. Podolske },
title = {Spaceborne detection of localized carbon dioxide sources},
journal = {Science},
volume = {358},
number = {6360},
pages = {eaam5782},
year = {2017},
doi = {10.1126/science.aam5782},
URL = {https://www.science.org/doi/abs/10.1126/science.aam5782},
eprint = {https://www.science.org/doi/pdf/10.1126/science.aam5782}}

@article{nassar2021,
title = {Advances in quantifying power plant {CO}2 emissions with {OCO}-2},
journal = {Remote Sensing of Environment},
volume = {264},
pages = {112579},
year = {2021},
issn = {0034-4257},
doi = {https://doi.org/10.1016/j.rse.2021.112579},
url = {https://www.sciencedirect.com/science/article/pii/S0034425721002996},
author = {Ray Nassar and Jon-Paul Mastrogiacomo and William Bateman-Hemphill and Callum McCracken and Cameron G. MacDonald and Tim Hill and Christopher W. O'Dell and Matthäus Kiel and David Crisp}
}

@article{shimodaira2000,
title = {Improving predictive inference under covariate shift by weighting the log-likelihood function},
journal = {Journal of Statistical Planning and Inference},
volume = {90},
number = {2},
pages = {227-244},
year = {2000},
issn = {0378-3758},
doi = {https://doi.org/10.1016/S0378-3758(00)00115-4},
url = {https://www.sciencedirect.com/science/article/pii/S0378375800001154},
author = {Hidetoshi Shimodaira}
}

@article{steininger2021,
  title={Density-based weighting for imbalanced regression},
  author={Steininger, Michael and Kobs, Konstantin and Davidson, Padraig and Krause, Anna and Hotho, Andreas},
  journal={Machine Learning},
  volume={110},
  number={8},
  pages={2187--2211},
  year={2021},
  publisher={Springer}
}

@inproceedings{yang2021,
  author={Yuzhe Yang and Kaiwen Zha and Ying-Cong Chen and Hao Wang and Dina Katabi},
  title={Delving into Deep Imbalanced Regression},
  year={2021},
  cdate={1609459200000},
  pages={11842-11851},
  url={http://proceedings.mlr.press/v139/yang21m.html},
  booktitle={{ICML}}
}

@article{malmgren-hansen2019,
title = {Statistical retrieval of atmospheric profiles with deep convolutional neural networks},
journal = {ISPRS Journal of Photogrammetry and Remote Sensing},
volume = {158},
pages = {231-240},
year = {2019},
issn = {0924-2716},
doi = {https://doi.org/10.1016/j.isprsjprs.2019.10.002},
url = {https://www.sciencedirect.com/science/article/pii/S0924271619302394},
author = {David Malmgren-Hansen and Valero Laparra and Allan {Aasbjerg Nielsen} and Gustau Camps-Valls}
}

@article{fabbro2018,
  title={An application of deep learning in the analysis of stellar spectra},
  author={Fabbro, S and Venn, KA and O'Briain, T and Bialek, S and Kielty, CL and Jahandar, F and Monty, S},
  journal={Monthly Notices of the Royal Astronomical Society},
  volume={475},
  number={3},
  pages={2978--2993},
  year={2018},
  publisher={Oxford University Press}
}

@article{guo2016entity,
  title={Entity embeddings of categorical variables},
  author={Guo, Cheng and Berkhahn, Felix},
  journal={arXiv preprint arXiv:1604.06737},
  year={2016}
}

@article{gorishniy2021,
  title={Revisiting deep learning models for tabular data},
  author={Gorishniy, Yury and Rubachev, Ivan and Khrulkov, Valentin and Babenko, Artem},
  journal={Advances in neural information processing systems},
  volume={34},
  pages={18932--18943},
  year={2021}
}

@article{vaswani2017attention,
  title={Attention is all you need},
  author={Vaswani, Ashish and Shazeer, Noam and Parmar, Niki and Uszkoreit, Jakob and Jones, Llion and Gomez, Aidan N and Kaiser, {\L}ukasz and Polosukhin, Illia},
  journal={Advances in neural information processing systems},
  volume={30},
  year={2017}
}

@article{nagrani2021attention,
  title={Attention bottlenecks for multimodal fusion},
  author={Nagrani, Arsha and Yang, Shan and Arnab, Anurag and Jansen, Aren and Schmid, Cordelia and Sun, Chen},
  journal={Advances in neural information processing systems},
  volume={34},
  pages={14200--14213},
  year={2021}
}

@article{bengio2013representation,
  title={Representation learning: A review and new perspectives},
  author={Bengio, Yoshua and Courville, Aaron and Vincent, Pascal},
  journal={IEEE transactions on pattern analysis and machine intelligence},
  volume={35},
  number={8},
  pages={1798--1828},
  year={2013},
  publisher={IEEE}
}

@article{yip2021,
doi = {10.3847/1538-3881/ac1744},
url = {https://doi.org/10.3847/1538-3881/ac1744},
year = {2021},
month = {oct},
publisher = {The American Astronomical Society},
volume = {162},
number = {5},
pages = {195},
author = {Yip, Kai Hou and Changeat, Quentin and Nikolaou, Nikolaos and Morvan, Mario and Edwards, Billy and Waldmann, Ingo P. and Tinetti, Giovanna},
title = {Peeking inside the Black Box: Interpreting Deep-learning Models for Exoplanet Atmospheric Retrievals},
journal = {The Astronomical Journal}
}

@article{rabanser2019failing,
  title={Failing loudly: An empirical study of methods for detecting dataset shift},
  author={Rabanser, Stephan and G{\"u}nnemann, Stephan and Lipton, Zachary},
  journal={Advances in Neural Information Processing Systems},
  volume={32},
  year={2019}
}

@InProceedings{schmitt2024detecting,
author="Schmitt, Marvin
and B{\"u}rkner, Paul-Christian
and K{\"o}the, Ullrich
and Radev, Stefan T.",
editor="K{\"o}the, Ullrich
and Rother, Carsten",
title="Detecting Model Misspecification in {Amortized Bayesian} Inference with Neural Networks",
booktitle="Pattern Recognition",
year="2024",
publisher="Springer Nature Switzerland",
address="Cham",
pages="541--557",
isbn="978-3-031-54605-1"
}

@article{gama2014,
  author  = {Gama, Jo{\~{a}}o and {\v{Z}}liobait{\.{e}}, Indr{\.{e}} and
             Bifet, Albert and Pechenizkiy, Mykola and Bouchachia, Abdelhamid},
  title   = {A survey on concept drift adaptation},
  journal = {ACM Computing Surveys},
  volume  = {46},
  number  = {4},
  pages   = {1--37},
  year    = {2014},
  doi     = {10.1145/2523813}
}

@article{parisi2019continual,
  title={Continual lifelong learning with neural networks: A review},
  author={Parisi, German I and Kemker, Ronald and Part, Jose L and Kanan, Christopher and Wermter, Stefan},
  journal={Neural networks},
  volume={113},
  pages={54--71},
  year={2019},
  publisher={Elsevier}
}

@article{
kirkpatrick2017overcoming,
author = {James Kirkpatrick  and Razvan Pascanu  and Neil Rabinowitz  and Joel Veness  and Guillaume Desjardins  and Andrei A. Rusu  and Kieran Milan  and John Quan  and Tiago Ramalho  and Agnieszka Grabska-Barwinska  and Demis Hassabis  and Claudia Clopath  and Dharshan Kumaran  and Raia Hadsell },
title = {Overcoming catastrophic forgetting in neural networks},
journal = {Proceedings of the National Academy of Sciences},
volume = {114},
number = {13},
pages = {3521-3526},
year = {2017},
doi = {10.1073/pnas.1611835114},
URL = {https://www.pnas.org/doi/abs/10.1073/pnas.1611835114},
eprint = {https://www.pnas.org/doi/pdf/10.1073/pnas.1611835114}}

@article{sculley2015hidden,
  title={Hidden technical debt in machine learning systems},
  author={Sculley, David and Holt, Gary and Golovin, Daniel and Davydov, Eugene and Phillips, Todd and Ebner, Dietmar and Chaudhary, Vinay and Young, Michael and Crespo, Jean-Francois and Dennison, Dan},
  journal={Advances in neural information processing systems},
  volume={28},
  year={2015}
}

@article{hammerling2012mapping,
  title={Mapping of {CO}2 at high spatiotemporal resolution using satellite observations: Global distributions from {OCO}-2},
  author={Hammerling, Dorit M and Michalak, Anna M and Kawa, S Randolph},
  journal={Journal of Geophysical Research: Atmospheres},
  volume={117},
  number={D6},
  year={2012},
  publisher={Wiley Online Library}
}

@Article{he2023machine,
AUTHOR = {He, Sicong and Yuan, Yanbin and Wang, Zihui and Luo, Lan and Zhang, Zili and Dong, Heng and Zhang, Chengfang},
TITLE = {Machine Learning Model-Based Estimation of {XCO}$_2$ with High Spatiotemporal Resolution in {China}},
JOURNAL = {Atmosphere},
VOLUME = {14},
YEAR = {2023},
NUMBER = {3},
ARTICLE-NUMBER = {436},
URL = {https://www.mdpi.com/2073-4433/14/3/436},
ISSN = {2073-4433},
DOI = {10.3390/atmos14030436}
}

@article{liu2024mapping,
title = {Mapping high-resolution {XCO}$_2$ concentrations in {China} from 2015 to 2020 based on spatiotemporal ensemble learning model},
journal = {Ecological Informatics},
volume = {83},
pages = {102806},
year = {2024},
issn = {1574-9541},
doi = {https://doi.org/10.1016/j.ecoinf.2024.102806},
url = {https://www.sciencedirect.com/science/article/pii/S1574954124003480},
author = {Weican Liu and Rong Li and Jun Cao and Congwu Huang and Fan Zhang and Meigen Zhang}
}

@article{antezanalopez2025global,
title = "Global Daily Column Average {CO}2 at 0.1° × 0.1° Spatial Resolution Integrating {OCO}-3, {GOSAT}, {CAMS} with {EOF} and Deep Learning",
author = "Antezana Lopez, Franz Pablo and Guanhua Zhou and Guifei Jing and Kai Zhang and Liangfu Chen and Lin Chen and Yumin Tan",
year = "2025",
month = dec,
doi = "10.1038/s41597-024-04135-w",
volume = "12",
journal = "Scientific Data",
issn = "2052-4463",
publisher = "Nature Research",
number = "1",
}

@article{hwang2026reconstructing,
author = {Soomin Hwang and Hyunyoung Choi and Yoojin Kang and Jungho Im},
title = {Reconstructing long-term (2003–2019) global high-resolution {XCO}$_2$: bridging observational gaps with machine learning},
journal = {GIScience \& Remote Sensing},
volume = {63},
number = {1},
pages = {2627042},
year = {2026},
publisher = {Taylor \& Francis},
doi = {10.1080/15481603.2026.2627042},


URL = { 
    
        https://doi.org/10.1080/15481603.2026.2627042
    
    

},
eprint = { 
    
        https://doi.org/10.1080/15481603.2026.2627042
    
    

}

}

@Article{reuter2025,
AUTHOR = {Reuter, M. and Hilker, M. and N{\"o}el, S. and Di Noia, A. and Weimer, M. and Schneising, O. and Buchwitz, M. and Bovensmann, H. and Burrows, J. P. and B{\"o}sch, H. and Lang, R.},
TITLE = {Retrieving the atmospheric concentrations of carbon dioxide and methane from the {European} {Copernicus} {CO2M} satellite mission using artificial neural networks},
JOURNAL = {Atmospheric Measurement Techniques},
VOLUME = {18},
YEAR = {2025},
NUMBER = {1},
PAGES = {241--264},
URL = {https://amt.copernicus.org/articles/18/241/2025/},
DOI = {10.5194/amt-18-241-2025}
}

@phdthesis{keely2025dissertation,
  title={Machine Learning Applications for Atmospheric Retrieval from Earth Observing Spectroscopy with Robust Uncertainty Quantification},
  author={Keely, William Ryan},
  school={University of Oklahoma},
  address={Norman, OK, USA},
  year={2025},
  url={https://hdl.handle.net/11244/341664}
}
\end{document}